%% file: vldb_main.tex
\pdfoutput=1

\documentclass[sigconf, nonacm]{acmart}
\input{Prefix.tex}

\newcommand\vldbdoi{XX.XX/XXX.XX}
\newcommand\vldbpages{XXX-XXX}
\newcommand\vldbvolume{14}
\newcommand\vldbissue{1}
\newcommand\vldbyear{2020}
\newcommand\vldbauthors{\authors}
\newcommand\vldbtitle{\shorttitle} 
\newcommand\vldbavailabilityurl{https://anonymous.4open.science/r/StreamLearning-1F77/}
\newcommand\vldbpagestyle{plain} 

\begin{document}

\title{StreamFP: Learnable Fingerprint-guided Data Selection for Efficient Stream Learning}

\author{Tongjun Shi$^{1,2}$, Shuhao Zhang$^1$, Binbin Chen$^2$, Bingsheng He$^3$}
\affiliation{%
  \institution{
    $^1$ National Engineering Research Center for Big DataTechnology and System\\
    Services Computing Technology and System Lab\\
    Cluster and Grid Computing Lab\\ 
    School of Computer Science and Technology\\ 
    Huazhong University of Science and Technology, Wuhan, 430074, China\\
    $^2$ Singapore University of Technology and Design    $^3$ National University of Singapore}
}



\input{Body/1_abstract}

\maketitle

\pagestyle{\vldbpagestyle}
\begingroup\small\noindent\raggedright\textbf{PVLDB Reference Format:}\\
\vldbauthors. \vldbtitle. PVLDB, \vldbvolume(\vldbissue): \vldbpages, \vldbyear.\\
\href{https://doi.org/\vldbdoi}{doi:\vldbdoi}
\endgroup
\begingroup
\renewcommand\thefootnote{}\footnote{\noindent
This work is licensed under the Creative Commons BY-NC-ND 4.0 International License. Visit \url{https://creativecommons.org/licenses/by-nc-nd/4.0/} to view a copy of this license. For any use beyond those covered by this license, obtain permission by emailing \href{mailto:info@vldb.org}{info@vldb.org}. Copyright is held by the owner/author(s). Publication rights licensed to the VLDB Endowment. \\
\raggedright Proceedings of the VLDB Endowment, Vol. \vldbvolume, No. \vldbissue\ %
ISSN 2150-8097. \\
\href{https://doi.org/\vldbdoi}{doi:\vldbdoi} \\
}\addtocounter{footnote}{-1}\endgroup

\ifdefempty{\vldbavailabilityurl}{}{
\vspace{.3cm}
\begingroup\small\noindent\raggedright\textbf{PVLDB Artifact Availability:}\\
The source code, data, and/or other artifacts have been made available at \url{https://github.com/intellistream/StreamLearning}.
\endgroup
}

\compact
\input{Body/2_Introduction}
\compact
\input{Body/3_Background}

\compact
\input{Body/4_Settings}
\compact
\input{Body/5_0_Methods}
\compact
\input{Body/5_1_Selections}

\input{Body/5_2_Buffer}
\input{Body/5_3_Attunement}
\input{Body/6_Results}
\compact
\input{Body/7_RelatedWork}
\compact
\input{Body/8_Conclusion}


\bibliographystyle{ACM-Reference-Format}
\bibliography{oclbib}


\appendix
\input{Body/Appendix}

\end{document}

%% file: Prefix.tex
\usepackage[utf8]{inputenc} 
\usepackage[T1]{fontenc}    
\usepackage{hyperref}       
\usepackage{url}            
\usepackage{booktabs}       
\usepackage{amsfonts}       
\usepackage{nicefrac}       
\usepackage{microtype}      
\usepackage{xcolor}         
\hypersetup{
    colorlinks,
    linkcolor={red!50!black},
    citecolor={blue!50!black},
    urlcolor={blue!80!black}
}
\usepackage{graphicx}
\usepackage{xspace}
\usepackage{multirow}

\usepackage{comment}

\usepackage{subcaption}
\setcitestyle{numbers,square,comma}
\usepackage{url}
\usepackage{xspace}
\usepackage{amsmath,amsfonts}
\usepackage{commath}
\usepackage{graphicx}
\usepackage{subcaption}
\usepackage{algorithm}
\usepackage{algpseudocode}
\usepackage{multirow}
\usepackage{makecell}
\algrenewcomment[1]{\(\triangleright\) #1}
\algnewcommand{\LineComment}[1]{\State \(\triangleright\) #1}

\usepackage{pgfplots}
\pgfplotsset{compat=newest}

\usepackage{hyphenat}
\usepackage{xcolor} 

\theoremstyle{plain}
\newtheorem{theorem}{Theorem}
\newtheorem{lemma}{Lemma}     
\newtheorem{definition}{Definition}  

\newcommand{\system}{{\textsl{StreamFP}}\xspace}

\usepackage[para]{threeparttable}

\usepackage{colortbl}
\definecolor{lightgray}{gray}{0.9}

\newcommand{\white}[1]{\textcolor{white}{#1}}

\usepackage{booktabs}
\usepackage{pifont}
\newcommand{\cmark}{\ding{51}}%
\newcommand{\xmark}{\ding{55}}%

\usepackage{enumitem}
\newenvironment{myitemize}
{ \begin{itemize}[leftmargin=0.2in]	
		\setlength{\itemsep}{0pt}
		\setlength{\parskip}{0pt}
		\setlength{\parsep}{0pt}    }
	{ 	 \end{itemize}                    }

\newenvironment{myenumerate}
{ \begin{enumerate}[leftmargin=0.2in]
		\setlength{\itemsep}{0pt}
		\setlength{\parskip}{0pt}
		\setlength{\parsep}{0pt}    }
	{ \end{enumerate}                  }

\newcommand{\compact}{\vspace{-2pt}} 
 
\usepackage[skip=7pt]{caption}

\usepackage{tikz}

\definecolor{green}{RGB}{99, 145, 122}
\newcommand{\codecomment}[1]{\textcolor{green}{\texttt{#1}}}


%% file: Body/1_abstract.tex
\begin{abstract} 
Stream Learning (SL) requires models that can quickly adapt to continuously evolving data, posing significant challenges in both computational efficiency and learning accuracy. 
Effective data selection is critical in SL to ensure a balance between information retention and training efficiency. Traditional rule-based data selection methods struggle to accommodate the dynamic nature of streaming data, highlighting the necessity for innovative solutions that effectively address these challenges. Recent approaches to handling changing data distributions face challenges that limit their effectiveness in fast-paced environments. In response, we propose \system, a novel approach that uniquely employs dynamic, learnable parameters called \textit{fingerprints} to enhance data selection efficiency and adaptability in stream learning. \system optimizes coreset selection through its unique fingerprint-guided mechanism for efficient training while ensuring robust buffer updates that adaptively respond to data dynamics, setting it apart from existing methods in stream learning. Experimental results demonstrate that \system outperforms state-of-the-art methods by achieving accuracy improvements of 15.99\%, 29.65\%, and 51.24\% compared to baseline models across varying data arrival rates, alongside a training throughput increase of 4.6x.
\end{abstract}

%% file: Body/2_Introduction.tex
\section{Introduction}
\label{sec:intro}
Stream learning (SL), the ability of models to adapt to continuously arriving, large-scale data streams, is crucial for applications like online recommendation systems~\cite{huang2015tencentrec} and autonomous driving~\cite{teichmann2018multinet}. This paradigm faces substantial challenges in balancing computational efficiency with learning accuracy. Major platforms demonstrate the scale of these challenges: Facebook processes 14.58 million photos hourly and Alibaba handles 583,000 transactions per second during peak periods~\cite{li2022camel}. In such high-throughput environments, rapidly evolving data distributions require models to integrate new information while preserving existing knowledge in real time. 
Data selection plays a crucial role in SL, significantly impacting model's adaption in streaming data through two key mechanisms: \textbf{coreset selection} and \textbf{buffer updates}. \textit{Coreset selection} identifies and selects informative training samples from each streaming data batch to adapt to evolving data distributions in high-volume streams.
As demonstrated by Camel~\cite{li2022camel}, it employs submodular maximization as the objective function to select representative data subsets.
\textit{Buffer updates} address catastrophic forgetting through proactive optimization that balances performance on new tasks while maintaining previously acquired capabilities. This approach differs fundamentally from concept drift detection~\cite{wen2024onenet, lu2018learning} in streaming scenarios, which reactively triggers model adaptation upon detecting novel distributions.
Rehearsal buffers mitigate forgetting by maintaining a subset of training data~\cite{borsos2020coresets}, with buffer updates determining data retention and replacement policies when new samples exceed capacity constraints. 
For instance, ASER~\cite{shim2021online} leverages Shapley Values to optimize memory updates from incoming data streams.
Current data selection methods in SL primarily comprise \textit{rule-based} and \textit{model-based} approaches. Rule-based methods employ predefined strategies to select data subsets~\cite{sener2017active, mirzasoleiman2020coresets, li2022camel, xie2024towards}. In contrast, model-based approaches utilize dynamic feedback mechanisms, such as gradients or losses, to evaluate data importance~\cite{yoo2019learning, killamsetty2021grad}. Despite their respective advantages, both approaches face challenges in optimizing the trade-off between computational efficiency and learning effectiveness in SL. We identify three critical limitations:



\textbf{Issue 1) Suboptimal Coreset Selection:}
While rule-based methods~\cite{sener2017active, mirzasoleiman2020coresets, li2022camel, xie2024towards} perform well in static environments, they struggle to adapt to streaming data's evolving nature, leading to outdated coresets that fail to capture recent patterns. Model-based approaches address this limitation by leveraging real-time model states to reflect changing data distributions. However, these approaches incur significant computational overhead, with increasing processing times that make them impractical for real-time analysis in high-throughput streaming environments.

\textbf{Issue 2) Unreliable Buffer Update:}
In SL, managing catastrophic forgetting requires a well-maintained rehearsal buffer that retains past knowledge. Rule-based methods like ER~\cite{chaudhry2019tiny}, ASER~\cite{shim2021online}, and Camel~\cite{li2022camel} update the buffer without the reliance on model states, reducing model involvement costs. However, these methods cannot ensure alignment with the evolving model states as the data stream progresses. Model-based approaches, such as GSS~\cite{aljundi2019gradient} and SSD~\cite{gu2024summarizing}, synchronize buffer updates with model evolution by leveraging gradients. Despite their dynamic updates, these methods introduce significant computational overhead, impairing real-time processing capabilities. 

\textbf{Issue 3) Unable to Adapt to Distribution Drifts:}
Distribution drift further complicates coreset selection and buffer updates. Rule-based methods struggle to adapt to varying data distributions without increasingly complex calculations, while model-based methods rely on additional network modules to enhance the robustness of model states (e.g., gradients or losses). These optimizations incur substantial computational overhead during training, adversely affecting the model's real-time performance and requiring efficient mechanisms to facilitate rapid model updates while enhancing state quality under distribution drifts.




Building an effective stream learning system requires solving challenges in data selection, particularly for coreset construction and buffer updates. Our approach introduces learnable parameters, called \textit{fingerprints}, to optimize these processes and improve adaptability and efficiency.
We use Vision Transformers (ViT)~\cite{dosovitskiy2020image} as an example due to their popularity in streaming applications such as video analytics~\cite{chen2022mm} and autonomous driving~\cite{prakash2021multi, shao2023safety}. Besides, our method can generalize to other transformer architectures, including CLIP~\cite{radford2021learning} and LLaMA~\cite{touvron2023llama}, with minimal adjustments. Unlike earlier works~\cite{wang2022learning, smith2023coda} that focus on model performance alone, our framework combines fingerprints with optimized data selection to achieve both adaptability and efficiency in stream learning.
To realize this vision, we must address the technical challenges that hinder effective stream learning: ensuring dynamic coreset selection, optimizing buffer updates, and refining fingerprints adjustments in real time.
\begin{myitemize}
\item 
\textbf{Suboptimal Coreset Selection:} The first challenge is to continuously select the most relevant subset of data—the coreset—for training. As data distributions shift over time, fingerprints must dynamically identify and prioritize data that reflects these changes. This ensures that the model stays aligned with the most current trends in the data, enhancing its ability to generalize and adapt effectively.
\item 
\textbf{Unreliable Buffer Updates:} Once the coreset is selected, the next challenge is managing the rehearsal buffer, where past data is stored to prevent catastrophic forgetting. Fingerprints should orchestrate the buffer update mechanism by identifying salient historical samples, thus enabling the model to preserve crucial knowledge while assimilating incoming data streams. This critical balance underpins sustained model performance.
\item 
\textbf{Inefficient Fingerprint Optimization:} The effectiveness of both coreset selection and buffer updates hinges on the continuous optimization of fingerprints. In high-throughput streaming environments, fingerprint optimization must happen in real time to keep up with the rapid flow of data. The entire system's performance degrades if fingerprint adjustments lag behind. Therefore, it is essential that fingerprints are fine-tuned efficiently and in sync with the evolving data stream, ensuring that the model remains responsive and accurate.
\end{myitemize}

In response to data selection challenges in stream learning (SL), we propose \system, a method that enables the model's real-time adaption through dynamic, learnable fingerprints. \system contains fingerprint-based coreset selection and buffer updates, along with continuous fingerprint attunement.

\system treats data relevance as an evolving attribute, using dynamic fingerprints to adaptively optimize both coreset selection and rehearsal buffer updates. This approach accelerates model training and ensures that the model remains responsive to the most pertinent data, even as the data stream evolves. As fingerprints are continuously optimized alongside the model, they directly influence coreset selection and buffer updates, allowing the system to adapt efficiently to rapidly changing data scenarios and mitigate catastrophic forgetting. To enable real-time fingerprint optimization, \system introduces fingerprint attunement, which leverages pretrained ViT attention layers through an optimized gating unit. This design achieves efficient fingerprint refinement with minimal computational overhead, making it scalable for continuous data streams.
In summary, we make following contributions:
\begin{myitemize}
\item We introduce \system, a novel stream learning (SL) framework that achieves real-time transformer model adaption in streaming environments. Introducing dynamic, learnable fingerprints, \system combines the coreset selection and buffer update to process high-throughput streams while guaranteeing both computational efficiency and learning quality.
\item We introduce compact, learnable fingerprints with a novel attunement mechanism based on pretrained Vision Transformer attention layers, enabling efficient model state adaptation with minimal overhead.
\item Extensive experiments across real-world datasets (Clear10, Clear100, CORe50, Stream-51) show \system achieves 15.99-51.24\% higher accuracy and 4.6x faster training throughput compared to state-of-the-art methods under varying data arrival rates. Implementation available at \url{https://github.com/intellistream/StreamLearning}.
\end{myitemize}
The remainder of this paper is structured as follows: Sec.~\ref{sec:ps} defines our technical challenges; Sec.~\ref{sec:methodology} presents \system's architecture; Sec.~\ref{sec:pcs}, Sec.~\ref{sec:pbm}, and Sec.~\ref{sec:pa} detail coreset selection, buffer updates, and fingerprint refinement respectively; Sec.~\ref{sec:exp} presents experimental results; Sec.~\ref{sec:related} reviews related work; and Sec.~\ref{sec:conclusion} concludes this paper.

%% file: Body/3_Background.tex
\section{Problem Statement}
\label{sec:ps}
\subsection{Preliminaries}

\begin{definition} \label{def:sl}
\textbf{(Stream Learning Problem)}  
Consider a stream-based classification problem where data arrives in sequential batches at a high rate $\lambda$ (samples/sec). Besides, consider a parameter space $\Theta$, a loss function $\mathcal{L}(\cdot,\cdot; \theta)$ parametrized by $\theta \in \Theta$ and a model update function $f(\cdot)$. At each timestamp $t$, let $B^t = \{(x_i, y_i)\}_{i=1}^{b}$ be a batch of instances, where $x_i \in \mathbb{R}^d$ represents the feature vector and $y_i \in \mathbb{R}^K$ denotes the corresponding label.
Given a new batch $B^t$ and a transformer model $\theta^{t-1}$ trained on previous batches, with the learning velocity $\mathcal{V}_{\theta}$ samples/sec, the stream learning problem aims to update the model parameters to $\theta^t$ while maintaining a processing speed that matches the arrival rate $\lambda$, thereby minimizing the accumulation of unprocessed data. The problem can be mathematically formulated as below:
\begin{align*}
    &\min_{\theta^t \in \Theta} \frac{1}{|B^t|}\sum_{(x_{i}, y_{i}) \in B^t}\mathcal{L}(x_{i},y_{i}; \theta^t)\\
    &\text{s.t.} \left\{
    \begin{array}{l}
        \theta^t = f(\theta^{t-1}, B^t)\\
        (\mathcal{V}_{\theta} - \lambda)^2 \to \min, \quad \mathcal{V}_{\theta} \gg \lambda
    \end{array}
    \right.
\end{align*}
\end{definition}
Note that the second constraint on learning velocity matches model training with data arrival rate to avoid both knowledge lag from insufficient training.

In this work, the transformer model must generalize across data with rapid shifting distributions, highlighting the importance of careful data selection and efficient model updates. The gradient-based methods updating the model $\theta^{t-1}$ can be expressed mathematically as:
\begin{equation}    \label{eq:gd}
\theta^{t,k} = \theta^{t,k-1} - \eta \frac{1}{|B^{t}|} \sum_{(x_{i}, y_{i}) \in B^{t}} \nabla_{\theta} \mathcal{L}(x_{i},y_{i}; \theta^{t, k-1}), 
\end{equation}
where $\theta^{t, 0} = \theta^{t-1, K}$ and $k \in \{1,2,\ldots, K\}$ is the number of gradient descent steps applied in each timestamp $t$.
$\eta$ is the learning rate, and $\nabla_{\theta}$ denotes the gradient with respect to parameters $\theta$.

\begin{definition} \label{def:pcl}
\textbf{(Fingerprint-based Continual Learning)} 
Consider $P \in \mathbb{R}^{N \times L_p \times D}$ denote a set of learnable parameters, i.e., \textbf{fingerprints}, where $N$ represents the number of fingerprints, $L_p$ denotes the fingerprint length, and $D$ is the embedding dimension. These fingerprints are prepended to specified Transformer layers, participating in forward propagation and being optimized through backward gradient computation. 
Given a new batch $B^t$ and a pretrained transformer model $\theta^{t-1}$, Fingerprint-based Continual Learning (FCL)~\cite{wang2022learning, wang2022s, smith2023coda}\footnote{It is also termed as \textit{Prompt for Continual Learning.}} aims to insert fingerprints to the model (i.e., $P^{t-1} \in \theta^{t-1}$) and only optimize the them from $P^{t-1}$ to $P^t$ while maintaining the frozen state of the original model parameters~\cite{smith2023coda}.
\end{definition}
This approach enables adaptation to novel distributions by utilizing the optimized fingerprints for inference guidance while preserving the pretrained knowledge in the frozen model parameters. 
FCL traditionally focuses on refining these fingerprints to improve continual learning (CL) performance across sequential data. Specifically, the fingerprints are integrated into the multi-head self-attention (MSA) mechanism by concatenating $\{p_K, p_V\} \in \mathbb{R}^{N\times \frac{L_p}{2} \times D}$ with key and value states:
\begin{equation} \label{eq:msa}
\begin{gathered}
h^{'}_K = p_K \cup h_K, \quad h^{'}_V = p_V \cup h_V, \\
\text{MSA}\left(h_Q, h^{'}_K, h^{'}_V\right) = (h_1 \cup \ldots \cup h_m)W^O, \\
h_i = \text{Attention}\left(h_Q W_i^Q, h^{'}_K W_i^K, h^{'}_V W_i^V\right),
\end{gathered}\nonumber
\end{equation}
where $\{h_K, h_V\}$ is the original key and value states in each attention layer, $W^O$, $W^Q$, $W^K$, and $W^V$ are projection matrices, and $m$ is the number of heads. Only $\{p_K, p_V\}$ are learnable, allowing the fingerprints to adapt as new data is processed.

In \system, the fingerprint update mechanism is equivalent to $k$-step gradient descent (GD). While a single step may appear insufficient, its computational efficiency enables rapid updates and swift adaptation to ever-changing distributions, making it effective for processing non-stationary data streams that demand real-time responses. These fingerprints are leveraged not just for model adaptation but also for managing data selection and model updates efficiently. This approach is further elaborated in the subsequent section, where we discuss how fingerprints are used for dynamic coreset selection and buffer update in stream learning.

%% file: Body/4_Settings.tex

\subsection{Problem Definition}
\label{sec:setting}
\begin{definition} \label{def:cs}
\textbf{(Coreset Selection for Stream Learning)}
Consider the stream learning problem in Def.~\ref{def:sl}, where data arrives sequentially in batches. Given batch $B^t=\{(x_i, y_i)\}_{i=1}^{b}$ at time step $t$ with rate $\lambda$, we aim to select a coreset $C^t=\{(x_i, y_i)\}_{i=1}^{c}$ ($C^t \subseteq B^t$) that minimizes the approximation loss $\mathbf{L}(\cdot, \cdot;\theta)$ of model $\theta^{t,k-1}$ between training on the coreset versus the full batch, where the model update velocity $\mathcal{V_\theta}$ should match the data arrival rate $\lambda$:
\begin{equation}
    \min_{\theta^t \in \Theta}\mathbf{L}(B^t,C^t;\theta) = |\frac{1}{b}\mathcal{L}(B^t;\theta^t)- \frac{1}{c}\mathcal{L}(C^t;\theta^t)|. \nonumber
\end{equation}

\end{definition}
To reduce the entire training time to match $\lambda$, \system leverages a set of $N$ fingerprints $P \in \mathbb{R}^{N \times L_p \times D}$ to efficiently select the proper coreset that evolves with the data stream.

\begin{definition} \label{def:bu}
\textbf{(Buffer update for Stream Learning)}
Consider the stream learning problem in Def.~\ref{def:sl}, where catastrophic forgetting needs to be addressed through rehearsal memory. Given a memory buffer $M^{t-1}$ ($|M|=m$) that stores $m$ past samples for rehearsal, we aim to obtain a new buffer $M^{t}$ by selecting samples from $M^{t-1}$ and $B^t$ under the size constraint $m$.

\end{definition}

As the model evolves with the stream, \system updates the buffer by selecting appropriate new streaming data and removing old buffer samples. This update ensures that the buffer consistently represents both current and past for model rehearsal. During model training on the new stream, \system replays past samples by integrating $M^{t-1}$ and $C^t$ to update the fingerprints $P$ of model $\theta^{t-1}$.

\begin{definition} \label{def:pa}
\textbf{(Fingerprint Attunement for Stream Learning)}
Consider the fingerprint parameters $p$ in Def.~\ref{def:pcl}. Let $\theta_f(\cdot; B^t)$ be a feature refiner (e.g., multilayer perceptron (MLP)) within the frozen ViT model $\theta$ to refine the fingerprints $P$ over streaming batch data $B^t$. Given a new streaming batch $B^t$, we aim to train feature refiner $\theta_f(\cdot; B^t)$ to better refine fingerprint parameters $P$.
\end{definition}

To achieve this, \system leverages pretrained ViT knowledge through a lightweight refinement mechanism, enabling efficient fingerprint refinement during streaming learning.

%% file: Body/5_0_Methods.tex
\begin{figure*}
    \centering
    \includegraphics[width=0.75\linewidth]{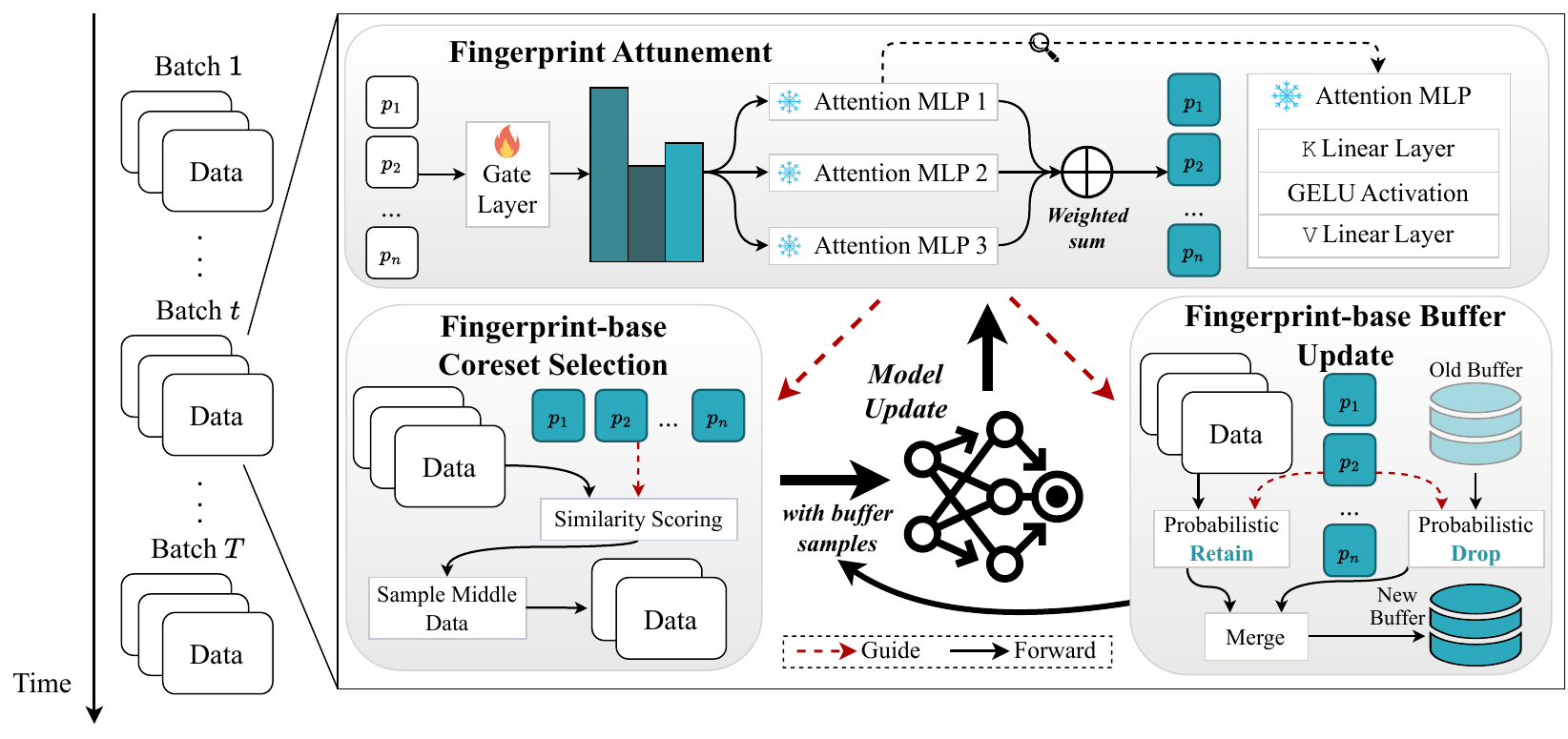}
    \caption{The overview of \system. Three components—Fingerprint-based Data Selection, Fingerprint-based Buffer Update, and Fingerprint Attunement—work synergistically to enhance training efficiency and preserve accuracy in stream learning.}
    \label{fig:method}
\end{figure*}

\section{System Design}
\label{sec:methodology}
Figure~\ref{fig:method} provides an overview of \system architecture, highlighting its key components and their interactions. Given the newly coming batch $B^t$, \system works in three stages: coreset selection, which identifies the most informative subset of data for training; model update
including fingerprint attunement, which refines the model's parameters; and buffer update, which manages the storage of historical data for effective learning.

\subsection{Stage 1: Coreset Selection}
This section details how \system efficiently selects representative samples from streaming batch data using fingerprint-sample similarity. The cornerstone of our methodology is a novel fingerprint-based coreset selection mechanism that achieves computational efficiency through the strategic utilization of fingerprints. Given an incoming streaming batch $B^t$, \system selects a coreset $C^t$ that captures essential information while reducing training overhead. The key challenge is to utilize the model's current state to select samples that reflect evolving data distributions, without incurring the computational cost of processing each sample through the entire model.

Our approach leverages fingerprints as the compact representations of the model state. These fingerprints evolve with the data stream, capturing the model's understanding of the current distribution. The process consists of two key steps: 1) extracting fingerprints from the current transformer model as dynamic templates for importance evaluation, and 2) computing similarity scores between these fingerprints and samples from batch $B^t$ to select samples with median-proximate similarities for inclusion in coreset $C^t$. This fingerprint-based methodology enables efficient importance evaluation through fingerprint-sample similarity computation with the related efficiency-quality guarantee (see Sec.~\ref{sec:pcs}).

\subsection{Stage 2: Model Update}
At this stage, the model of \system is updated by fingerprint-based continual learning while the fingerprints are further refined by our proposed fingerprint attunement.

\noindent\textbf{Fingerprint-based Continual Learning.}
Based on the coreset $C^t$ and the buffer $M^{t-1}$, \system updates the model in the fingerprint-based continual learning manner, i.e., only updating the fingerprint parameters while keeping the model parameters frozen. Specifically, given the newly coming batch $B^t$, \system firstly constructs the coreset $C^t$ as processed batch data $B_C$. Besides, \system obtains the buffer mini-batch $B_M$ by randomly retrieving the buffer $M^{t-1}$. Combining $B_C$ and $B_M$ as the training batch $B_{\text{train}}$, \system updates the model through the forward (Eq.\ref{eq:msa}) and the gradient descent until a single pass over $B_{\text{train}}$. After the model finishes the training on $C^t$, \system updates the buffer from $M^{t-1}$ to $M^{t}$ based on the whole batch $B^t$ for the next model rehearsal. 

\noindent\textbf{Fingerprint Attunement Mechanism.}
To effectively handle distribution drift in streaming scenarios, \system introduces a fingerprint attunement mechanism that addresses two critical challenges: 1) the need for effective optimization of model states to maintain high-quality coreset selection and buffer updates, and 2) the requirement for computational efficiency to preserve the benefits of these optimizations.

Our solution implements fingerprint refinement through a novel architecture combining frozen MLPs~\cite{tolstikhin2021mlp} with a learnable gate unit. This gate unit adaptively refines fingerprints by utilizing pretrained knowledge while requiring only minimal parameter updates. By focusing the optimization on these compact gate parameters, our mechanism achieves efficient fingerprint refinement (see Sec.~\ref{sec:pa}).

\subsection{Stage 3: Buffer Update}
To mitigate catastrophic forgetting, \system maintains a memory buffer that preserves training samples after processing each incoming mini-batch.
This step aims to store representative training data for subsequent model training rehearsal. Previous buffer update methods can be categorized as rule-based (e.g., ER~\cite{chaudhry2019tiny}, ASER~\cite{shim2021online}, and Camel~\cite{li2022camel}) or model-based (e.g., GSS~\cite{aljundi2019gradient} and SSD~\cite{gu2024summarizing}). However, rule-based methods are unable to handle evolving model states due to their static nature, while model-based methods introduce significant computational overhead, reducing real-time processing speed. Consequently, both approaches lack either effectiveness (rule-based methods) or efficiency (model-based methods) for model rehearsal.

To address these issues, we propose an efficient method by leveraging fingerprints for selecting training samples that align with the evolving characteristics of past data.
Specifically, we compute batch-fingerprint and buffer-fingerprint similarities, which we then use to derive a probability distribution. Subsequently, we employ probabilistic sampling to select new batch data and remove old buffer data. The utilization of fingerprints $P$ provides both computational efficiency and quality guarantees for buffer maintenance (Sec.~\ref{sec:pbm}).

%% file: Body/5_1_Selections.tex
\section{Fingerprint-based Coreset Selection}
\label{sec:pcs}
\begin{figure}
    \centering
    \includegraphics[width=1\linewidth]{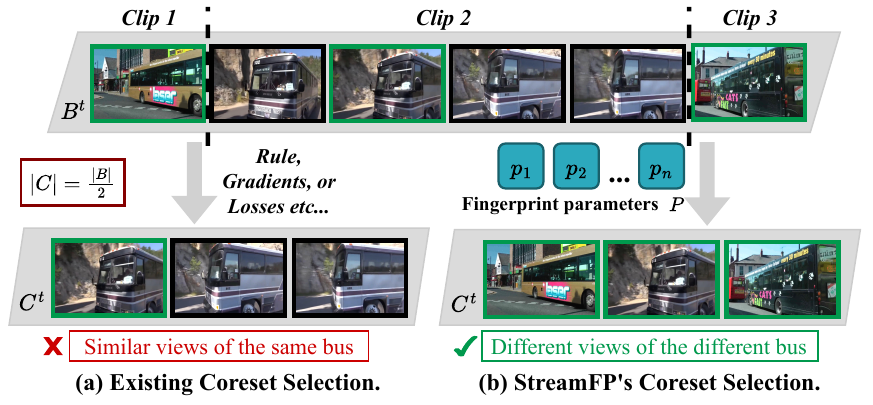}
    \caption{Comparison of existing work vs. our approach: Current methods bias data selection towards dominant clips in diverse streaming batches, reducing model generalizability. Our approach uses knowledge-inheriting fingerprints to create a diverse, representative coreset and buffer, enhancing model performance.}
    \label{fig:cs_comp}
\end{figure}
To enhance computational efficiency in stream learning, \system employs a sophisticated coreset selection strategy leveraging fingerprints to filter data effectively. As illustrated in Figure~\ref{fig:cs_comp}, unlike existing methods that tend to select redundant data points (Figure~\ref{fig:cs_comp}a), our fingerprint-based approach ensures the model focuses on the most relevant and informative data points by maintaining diversity in selection (Figure~\ref{fig:cs_comp}b). This approach improves overall learning efficiency and adaptability.
At the start of each training iteration, embeddings of the batch data $B^t$ are extracted using the first embedding layer of the model, denoted as $emd \in \mathbb{R}^{b \times L \times D}$, where $b$ is the batch size, $L$ is the sequence length, and $D$ is the dimension of the embeddings. 
Concurrently, for the extracted fingerprint parameters $P \in \mathbb{R}^{N \times L_p \times D}$, we aggregate the fingerprint length dimension (i.e., $\text{torch.sum}(P, dim=1)$) to obtain $P \in \mathbb{R}^{N \times 1 \times D}$ and reshape $P$ to $P \in \mathbb{R}^{N \times D}$. Here, $N$, and $L_p$ represent the number and length of the fingerprints respectively, where the fingerprints' dimension (768) aligns with the data embeddings for subsequent operations.
We compute the cosine similarity between the data embeddings and fingerprints to select new data $X_{new}$ using the following formula:
\begin{equation}
\label{eq:cos}
\begin{gathered}
S'=\text{sim}(emd, P) = \frac{P \cdot emd}{\|P\| \|emd\|},\\
S = \text{mean}(S').
\end{gathered}
\end{equation}

 Regarding the input of $S'$ where $P \in \mathbb{R}^{N \times D}$ and $emd \in \mathbb{R}^{b \times L \times D}$, we reshape $P$ to $\mathbb{R}^{1 \times D \times N}$ and normalize both $P$ and $emd$. We then utilize PyTorch's broadcasting mechanism and perform matrix multiplication to compute $S'=\text{torch.matmul}(emd, p)$, resulting in $S' \in \mathbb{R}^{b \times L \times N}$. To aggregate all fingerprints and sequence information into a single data point, we subsequently perform the $\text{mean}(\cdot)$ operation that we utilize $\text{torch.mean}(S', dim=(1,2))$ to obtain a tensor of shape $\mathbb{R}^{b \times 1 \times 1}$. Finally, we reshape this result to acquire the similarity $S \in \mathbb{R}^{b}$ between the fingerprints and data embeddings for selection:
\begin{equation}
\label{eq:select}
\begin{gathered}
c = \sigma \times b, \\
I_{\text{sorted}} = \text{argsort}(S), \\
C^t = { B^t[I_{\text{sorted}}[ (\lfloor \frac{b}{2} \rfloor - \lfloor \frac{c}{2}\rfloor) : (\lfloor \frac{b}{2} \rfloor + \lfloor \frac{c}{2}\rfloor)]] },
\end{gathered}
\end{equation}
where $c$ is the size of coreset $C^t$ with the corresponding coreset ratio $\sigma$ among $B^t$. Determining the standpoint as $\frac{b}{2}$, we select data from the vicinity of the midpoint in the newly ranked batch data index, based on similarity $S$. This selection identifies data points that balance novelty and familiarity, allowing the model to incorporate new knowledge while maintaining consistency with established learning. 
As demonstrated in the Supplementary A.1, the minimal correlation in fingerprint gradients across diverse task data reveals distinct optimization trajectories, evidencing that each task induces the fingerprint to acquire task-specific representations. Through direct extraction and implicit gradient reflection, fingerprints $P$ encapsulating model-specific knowledge enable the selection to effectively adapt to distribution drift. Consequently, this fingerprint-based selection enhances \system's robustness and efficiency in handling evolving data streams.

\textbf{Efficiency Guarantees.}
The space complexity of our algorithm is $\mathcal{O}(|P|)$, attributed to fingerprint storage requirements, while its theoretical time complexity is $\mathcal{O}(|B|\cdot|P|)$ due to single matrix multiplication in similarity computations. Through GPU acceleration, the effective runtime complexity reduces to $\mathcal{O}(\lceil (|B|\cdot|P|)/\gamma \rceil)$, where $\gamma$ denotes the GPU's parallel processing capacity, achieved via CUDA-optimized batch operations.

\textbf{Quality Guarantees.} We demonstrate that our fingerprint-based coreset selection method upholds the quality guarantees outlined in Theorem~\ref{theorem:cqg}, as detailed in the Supplementary A.3. 
This theorem provides theoretical validation for our fingerprint-based coreset selection strategy by establishing strong probabilistic guarantees. Specifically, it proves that the selected coreset $C^t$ preserves the average angular distance to fingerprints $P$ within a multiplicative error bound of $(1\pm\varepsilon)$ relative to the original batch $B^t$'s cost, where $\varepsilon = O(\sqrt{\log(1/\delta)/(\sigma b)})$. This error bound demonstrates that the approximation quality improves with larger batch sizes $b$ and coreset ratios $\sigma$. Moreover, the theorem's reliance on angular distance $\arccos(\text{sim}(x,P))$ naturally aligns with our cosine similarity-based selection mechanism. These theoretical guarantees validate that our selection strategy not only achieves computational efficiency but also maintains essential geometric relationships between data points and fingerprints, thereby ensuring robust knowledge preservation and adaptation in streaming scenarios.
\begin{theorem}[Coreset Quality Guarantee] \label{theorem:cqg}
With probability at least $1-\delta$, the coreset $C^t$ satisfies:
\[(1-\varepsilon)\text{cost}(B^t) \leq \text{cost}(C^t) \leq (1+\varepsilon)\text{cost}(B^t),\]
where $\text{cost}(X) = \frac{1}{|X|}\sum_{x\in X} d(x)$, representing the average angular distance, $d(x) = \arccos(\text{sim}(x,P))$, and $\varepsilon = O(\sqrt{\log(1/\delta)/(\sigma b)})$.
\end{theorem}

Our fingerprint-based coreset selection algorithm is shown in Alg.~\ref{alg:pcs}. The process initiates with data embedding extraction and fingerprint preprocessing (lines 1-5). Subsequently, it computes the similarity between data embeddings and processed fingerprints to facilitate further selection (lines 6-9). Finally, based on the resultant similarities, we select samples from the vicinity of the midpoint of the sorted list as the coreset (lines 10-13). The algorithm achieves high efficiency by eschewing loop statements and exclusively utilizing PyTorch's matrix multiplication operations.
\begin{algorithm}[t] 
\caption{Pseudocode of Fingerprint-based Coreset Selection.}
\label{alg:pcs}
\begin{algorithmic}[1]
\fontsize{8.75}{11}\selectfont
\Statex \textbf{Input:} Transformer model $\theta^{t-1}$, batch data $B^t = \{(x^t_i, y^t_i)\}_{i=1}^{b}$, batch size $b=|B|$, fingerprints parameters $P$, coreset ratio $\sigma$
\Statex \textbf{Output:} selected coreset $C^t$
\State \codecomment{// extract data embeddings}
\State $emd = \theta^{t-1}(B^t) \in \mathbb{R}^{b \times L \times D}$
\State \codecomment{// preprocess fingerprints parameters $P \in \mathbb{R}^{N \times L_p \times D}$}
\State $P_{n,d} = \sum_{l=1}^{L_p} P_{n,l,d}$
\State $P_{1,n,d} = P_{n,d}$ \codecomment{// $P \in \mathbb{R}^{1 \times N \times D}$}
\State \codecomment{// compute similarity according to Eq.\ref{eq:cos}}
\State $P_{1,n,d} = \frac{P_{1,n,d}}{\sqrt{\sum_{d=1}^D (P_{n,d})^2}}$, $emd_{b,l,d} = \frac{emd_{b,l,d}}{\sqrt{\sum_{d=1}^D (emd_{b,l,d})^{2}}}$ \codecomment{// L2 norm}
\State $S'_{b,l,n} = \sum_{d=1}^D emd_{b,l,d} \cdot P_{1,d,n}$ \codecomment{// $S' \in \mathbb{R}^{b \times L \times N}$}
\State $S_b = \frac{1}{L \cdot N} \sum_{l=1}^L \sum_{n=1}^N S'_{b,l,n}$ \codecomment{// $S \in \mathbb{R}^b$}
\State \codecomment{// select coreset $C^t$ according to Eq.\ref{eq:select}}
\State $c=\sigma \times b$
\State $I_{\text{sorted}} = \text{argsort}(S)$ \codecomment{// $S_{I_{\text{sorted}}[i]} \geq S_{I_{\text{sorted}}[i+1]}$}
\State $C^t=B^t[I_{\text{sorted}}[\lfloor \frac{b}{2} \rfloor - \lfloor \frac{c}{2} \rfloor:\lfloor \frac{b}{2} \rfloor+\lfloor \frac{c}{2} \rfloor]]$
\end{algorithmic}
\end{algorithm}

%% file: Body/5_2_Buffer.tex
\section{Fingerprint-based Buffer Update}
\label{sec:pbm}
Building on the enhanced efficiency from coreset selection, rehearsing data from the previous stream can mitigate catastrophic forgetting and improve model accuracy, as part of the overall
\system framework. Therefore, we equip the rehearsal buffer with \system, an effective update strategy that balances learning new information and retaining previously learned knowledge. Given that fingerprints evolve alongside the model, they can effectively reflect the model's states throughout the stream learning process. \system employs dynamic fingerprints to manage the rehearsal buffer update, ensuring it remains representative of the model's knowledge base while minimizing the maintaining overheads. As illustrated in Figure~\ref{fig:method}, \system performs buffer updates in a \textit{Retain-Drop} manner. Using probabilistic sampling based on the batch-fingerprint and buffer-fingerprint similarities allows for a more flexible approach to selectively retaining new data from the
streaming batch, dropping old data from the buffer, and merging the retained new data with the remaining buffer data to form the updated buffer. This approach ensures that the buffer remains representative for both current and past model states, effectively easing catastrophic forgetting in stream learning.

Firstly, we determine the number of buffer update samples $\nu^t$: 
\begin{equation}
\label{eq:upt_num}
n_{\text{left}} = b-\max(0, m-n_{\text{seen}}),
\end{equation}
where $\nu^t = \min(\lfloor\frac{b}{2}\rfloor, \max(1, |{\text{Sample}(n_{\text{left}}, \mathcal{U}(0, n_{\text{seen}})) < m}|))$.
In this formulation, $n_{\text{seen}}$ denotes the cumulative number of samples encountered in the buffer, and $\text{Sample}(n_{\text{left}}, \mathcal{U}(0, n_{\text{seen}}))$ represents sampling $n_{\text{left}}$ data points from a uniform distribution over the interval $[0, n_{\text{seen}})$. As $n_{\text{seen}}$ in the uniform distribution increases over time, the number of buffer update samples $\nu^t$ correspondingly decreases gradually. 

We then compute the batch-fingerprint and buffer-fingerprint similarities, denoted as $S_{\text{batch}}$ and $S_{\text{buffer}}$ respectively, utilizing Eq. \ref{eq:cos}. Subsequently, we apply relative rank probabilities to these similarity values. For a given similarity value $s_i \in S$ corresponding to the $i$-th data point, we determine its rank $r_i$ through descending order sorting, such that a lower $r_i$ corresponds to a higher $s_i$, and vice versa. The relative rank probabilities $\pi_i$ are calculated as follows:
\begin{equation}
\label{eq:buf_prob}
\pi_i = 1 - \frac{1/r_i}{\sum_{j=1}^n 1/r_j}, \
\text{where} \  r_i = \text{rank}(s_i) \  \text{and} \  s_i \geq s_j \iff r_i \leq r_j.
\end{equation}
Here, $\pi_i$ is inversely related to $s_i$, reflecting our objective to update the buffer with novel samples. By aggregating all $\pi_i$ values, we derive two probability distributions, $\pi_{\text{batch}}$ and $\pi_{\text{buffer}}$, corresponding to $S \in {S_{\text{batch}}, S_{\text{buffer}}}$. This approach enables us to prioritize the selection of samples that exhibit lower similarity to the existing data, thereby promoting diversity in the buffer.

Leveraging these probabilities as distributions, we implement a probabilistic sampling mechanism to update the buffer. This sampling preferentially retains batch and buffer data points with higher probabilities. The formal representation of this process is:
\begin{equation}
\label{eq:buf_update}
\begin{gathered}
I_{\text{batch}} = \mathbb{S}(B^t, \pi_{\text{batch}}, \nu^t),\\
I_{\text{buffer}} = \mathbb{S}(M^{t-1}, \pi_{\text{buffer}}, \nu^t),\\
\end{gathered}
\end{equation}
The function $\mathbb{S}(B^t, \pi, \nu^t)$ denotes probabilistic sampling, specifically the selection of $\nu^t$ data points from batch data $B^t$ according to the probability distribution $\pi$.
The buffer update process is completed by replacing the discarded buffer data with the retained batch data in their respective positions. This methodology ensures a dynamic and adaptive buffer that maintains relevance and diversity throughout the learning process.

\system remains efficient in managing the buffer due to two key factors: 1) the minimal cost of data embedding extraction, achieved by using only the model's first embedding layer rather than all layers; 2) the efficient computation of embedding similarity between data embeddings and fingerprints, owing to their relatively small number of parameters. Consequently, \system effectively maintains a cohesive and relevant dataset within the buffer by continually updating all buffer data.

\textbf{Efficiency-Quality Guarantee.} The time complexity is $\mathcal{O}(|B|+|M|)$. Furthermore, our buffer update method upholds the representativeness guarantee outlined in Theorem~\ref{theorem:burg} and its proof is detailed in the Supplementary A.4.
This theorem establishes theoretical guarantees for our fingerprint-based buffer update strategy by bounding the distributional difference between the maintained buffer and the full stream history. Specifically, it proves that the Maximum Mean Discrepancy (MMD) between the buffer distribution $P_M$ and the complete data distribution $P_t$ is bounded by $\varepsilon = \mathcal{O}((m\ln(1/\delta))^{1/4})$ with probability $(1-\delta)$. The kernel function $k(x,y) = \text{sim}(x,P)\text{sim}(y,P)$ in MMD naturally aligns with our similarity-based sampling mechanism, while the error bound's dependence on buffer size $m$ theoretically validates that larger buffers lead to better representation quality. These theoretical guarantees demonstrate that our buffer update strategy effectively maintains a representative subset of the streaming data, ensuring reliable knowledge preservation while managing memory constraints efficiently.
\begin{theorem}(Buffer Update Representativeness Guarantee) \label{theorem:burg}
With probability at least $1-\delta$, the distribution $P_M$ obtained from the buffer update satisfies:
\[
D(P_M, P_t)\leq \varepsilon,
\]
where $D(\cdot,\cdot)$ denotes the Maximum Mean Discrepancy (MMD) between distributions with kernel function $k(x,y) = \text{sim}(x,P)\text{sim}(y,P)$, $P_M$ is the distribution of buffer data, $P_t$ is the distribution of all seen data until time $t$, and $\varepsilon = \mathcal{O}((m\ln(1/\delta))^{1/4})$ with $m$ being the buffer size.
\end{theorem}

Our fingerprint-based buffer update algorithm is detailed in Algorithm \ref{alg:pbm}. Initially, we extract the embeddings of both batch and buffer data, and preprocess the fingerprint parameters before normalization (lines 1-9). Subsequently, we compute the batch-fingerprint and buffer-fingerprint similarities, applying the softmax function to derive the probability distribution (lines 10-14). Once the probability distribution is obtained, we employ probabilistic sampling to select new batch data and remove old buffer data, thereby updating the new buffer $M^t$ (lines 15-18). 
\begin{algorithm}[t] 
\caption{Pseudocode of Fingerprint-based Buffer Update.}
\label{alg:pbm}
\begin{algorithmic}[1]
\fontsize{8.75}{11}\selectfont
\Statex \textbf{Input:} current batch similarity $S_{\text{batch}}$, previous buffer similarity $S_{\text{buffer}}$, batch data $B^t = \{(x^t_i, y^t_i)\}_{i=1}^{b}$, batch size $b=|B|$, previous buffer $M^{t-1}=\{(x^{t-1}_i, y^{t-1}_i)\}_{i=1}^{m}$ buffer size $m=|M|$,, fingerprints parameters $P$, the number of data samples seen in the buffer $n_{\text{seen}}$
\Statex \textbf{Output:} updated buffer $M^t$
\State \codecomment{// determine the number of buffer updates by Eq.\ref{eq:upt_num}}
\State $n_{\text{left}} = b-\max(0, m-n_{\text{seen}})$
\State $\text{indices}_i \sim \mathcal{U}(0, n{\text{seen}})$ \codecomment{// $i = 1,2,...,n_{\text{left}}$}
\State $\nu^t = |{i : \text{indices}_i < m}|$
\State $ \nu^t = \min(\lfloor\frac{b}{2}\rfloor, \max(1, \nu^t))$
\State \codecomment{// compute probabilities according to Eq.\ref{eq:buf_prob}}
\State $I'_{\text{batch}} = \text{argsort}(S_{\text{batch}})$ \codecomment{// $S_{\text{batch}_{I_{\text{batch}}[i]}} \geq S_{\text{batch}_{I_{\text{batch}}[i+1]}}$}
\State $I'_{\text{buffer}} = \text{argsort}(S_{\text{buffer}})$ \codecomment{// $S_{\text{buffer}_{I_{\text{buffer}}[i]}} \geq S_{\text{buffer}_{I_{\text{buffer}}[i+1]}}$}
\State $r_{\text{batch}} = \{1, 2, ..., b\}$, $r_{\text{buffer}} = \{1, 2, ..., m\}$
\State $p_{\text{batch}} = 1 / r_{\text{batch}}$, $p_{\text{buffer}} = 1 / r_{\text{buffer}}$
\State $\pi_{\text{batch}} = 1 - p_{\text{batch}} / \sum p_{\text{batch}}$, $\pi_{\text{buffer}} = 1 - p_{\text{buffer}} / \sum p_{\text{buffer}}$
\State $\pi_{\text{batch}}[I'_{\text{batch}}] = \pi_{\text{batch}}$, $\pi_{\text{buffer}}[I'_{\text{buffer}}] = \pi_{\text{buffer}}$
\State \codecomment{// retain and drop data by Eq.\ref{eq:buf_update}}
\State $I_{\text{batch}}=\mathbb{S}(B^t, \pi_{\text{batch}}, \nu^t)$, $I_{\text{buffer}}=\mathbb{S}(M^{t-1}, 1 - \pi_{\text{buffer}}, \nu^t)$
\State \codecomment{// merge the resultant data points}
\State $M^t = (M^{t-1} \setminus M^{t-1}[I_{\text{buffer}}]) \cup B^{t}[I_{\text{batch}}]$
\end{algorithmic}
\end{algorithm}

%% file: Body/5_3_Attunement.tex
\section{Fingerprint Attunement}
\label{sec:pa}
Fingerprints play a crucial role in the coreset selection and buffer updates of \system. However, in stream learning, single-pass data processing hinders fingerprint learning, posing a significant challenge. Inspired by the robustness and general knowledge of ViTs, we propose \textit{fingerprint attunement} to fully exploit the expertise of attention layers in ViTs and fine-tune the initial fingerprints, as illustrated in Figure~\ref{fig:method}. The fingerprint attunement (i.e., the fingerprint feature refiner $\theta_f$ as illustrated in Def.~\ref{def:pa}) consists of two main components: a gate layer that dynamically adjusts weights based on relevance, and three attention MLPs that process these fingerprints.

Given the fingerprints $P$, we feed them into the gate layer. In the $l$-th layer of the ViT, the weights of the fingerprints for the $r$-th attention MLP outputs in fingerprint attunement are computed as follows:
\begin{equation} \label{eq:gate}
\begin{gathered}
(V, I) = \text{Top-r}\left((W_g \cdot P), R\right), \\
W = \text{softmax}(V),
\end{gathered}
\end{equation}
where $W_g \in \mathbb{R}^{D \times R}$ represents the weights of the gate layer, with $D = 768$ being the hidden dimension size of each attention MLP, and $R = 3$ being the number of total attention MLPs. Using the gate layer features, we select their top-3 values $V$ and corresponding indexes $I$, which then pass through the softmax function to become the final weights $W$ of the attention MLP features.

The core of fingerprint attunement is the attention MLP that fully exploits the expertise of ViT's attention layers. We propose to split the last three pretrained ViT attention layers into the key and value weights and load them into the linear layer of the attention MLP as the K and V linear layers, respectively. To maintain the general knowledge of preloaded attention weights, we freeze the updates of attention MLPs to avoid biasing towards specific distributions. Specifically, we reorder the fingerprints $P$ based on the gate index $I$, and the reordered outputs are fed into three distinct attention-based MLPs. Formally, the rest of fingerprint attunement, including the attention MLPs and output linear layer, is calculated as:
\begin{equation} \label{eq:attn_mlp}
\begin{gathered}
f_{attn_r} = W_{\texttt{V}_r} \cdot \text{GELU}\left(W_{\texttt{K}_r} \cdot \text{reorder}(p, I) \right),\\
f_{attn} = \text{concatenate}(f_{attn_1}, ..., f_{attn_R}),\\
p_{out} = W \cdot f_{attn},
\end{gathered}
\end{equation}
where $W_{K_r} \in \mathbb{R}^{D \times D}$ and $W_{V_r} \in \mathbb{R}^{D \times D}$ are the weights of K and V linear layers in the $r$-th attention MLP. Combining all MLP outputs as attention MLP features $f_{attn} \in \mathbb{R}^{N \times R \times D}$, we weight and sum the attention features by the feature weights $W \in \mathbb{R}^{N \times 1 \times R}$ obtained from the gate layer to get the final refined fingerprints $p_{out} \in \mathbb{R}^{N \times L_p \times D}$.


Our fingerprint attunement method improves upon prior approaches that use unrefined fingerprint parameters. While it inherits knowledge from pretrained models (e.g., ImageNet), adapting to streaming data typically requires costly fine-tuning that impedes real-time processing. To address this, we introduce a lightweight learnable gate unit that adaptively leverages ViT attention layers without extensive fine-tuning, enabling efficient single-pass adaptation to streaming data.

%% file: Body/6_Results.tex
\section{Experiments}
\label{sec:exp}
In this section, we present the results of our empirical evaluation, comparing our method against several closely related coreset selection baselines across different experimental conditions. The experiments were conducted on a server equipped with an i7-13700K CPU, a GeForce RTX A6000 GPU, and 64 GB of memory.

%
\subsection{Methodologies}
\label{experiment:setting}

\textbf{Datasets.} To evaluate the performance of data selection in a streaming environment, we use four datasets: Clear10, Clear100~\cite{lin2021clear}, CORe50~\cite{lomonaco2017CORe50}, and Stream-51~\cite{roady2020stream} which is summarizing in Table~\ref{tab:dataset}.
\begin{table*}[t]
\centering
\caption{Comparison of four streaming datasets.}
\label{tab:dataset}
\fontsize{15}{16}\selectfont
\resizebox{0.95\textwidth}{!}{%
\begin{tabular}{l|r|r|c|l}
\toprule
\multicolumn{1}{c|}{\textbf{Dataset}} & \multicolumn{1}{c|}{\textbf{Total Images}} & \multicolumn{1}{c|}{\textbf{Total Classes}} & \multicolumn{1}{c|}{\textbf{Distribution Drift Characteristics}} & \multicolumn{1}{c}{\textbf{Key Features}} \\
\midrule
Clear10 & 396,550 & 11 & Natural temporal evolution over a decade (2004-2014) & \makecell[l]{- Long-term concept evolution\\- Real-world temporal changes} \\
\midrule
Clear100 & 1,814,197 & 100 & Similar to Clear10, temporal   evolution across years & \makecell[l]{- Largest dataset in the collection\\- Balanced and comprehensive task distribution\\} \\
\midrule
CORe50 & 164,866 & 50 & Fine-grained temporal variations   within video sessions & \makecell[l]{- Objects in gentle motion\\- Intra-class temporal coherence\\-   Emphasis on stream sequencing} \\
\midrule
Stream-51 & 150,736 & 51 & Continuous video stream variations & \makecell[l]{- Fine-grained temporal   dynamics\\- Real-time data variations\\- Video-based   continuous stream} \\
\bottomrule
\end{tabular}}
\end{table*}

\textbf{Evaluation Metrics.}
We assess the performance of SL using Average Accuracy and Average Forgetting metrics~\cite{shim2021online}. Average Accuracy is calculated across all seen tasks and is defined as $\text{Average Accuracy} = \frac{1}{T} \sum_{j=1}^{T} a_{T,j}$, where $a_{i,j}$ denotes the accuracy on task $j$ after training from task 1 through $i$. Average Forgetting measures the extent of memory loss for each task following training on the final task and is defined as $\text{Average Forgetting} = \frac{1}{T-1} \sum_{j=1}^{T-1} f_{T,j}$, where $f_{i,j} = \max_{k \in {1,\ldots,i-1}} a_{k,j} - a_{i,j}$.

\textbf{Data Arrival Rates.}
As mentioned in Section~\ref{sec:setting}, real-world stream data does not always arrive at a rate that matches the model's learning speed~\cite{ghunaim2023real}. Following prior work~\cite{ghunaim2023real}, we simulate diverse arrival rates by skipping batches. Specifically, given an arrival rate and the dataset size, we calculate the overall duration during which all data arrives, referred to as the ``total duration.'' For each method, we run 500 batches to measure the training time per batch, then multiply the training time by the total number of batches to determine the expected overall training time. By dividing the expected overall training time by the total duration, we obtain the stream-model relative complexity $C_S=\frac{\text{expected overall training time}}{\text{total duration}}$. When $C_S > 1$, data has to be skipped as training time exceeds the total duration. We then perform uniform sampling across all batch data, retaining only a fraction $\frac{1}{C_S}$ of the batches, effectively skipping batches. Unless mentioned otherwise, for coreset selection methods, we apply a selection ratio $\sigma$ of 0.5, selecting half of the data from each non-skipped batch. For buffer-related methods, we maintain a rehearsal buffer with a size $M$ of 102. The impact of varying $\sigma$ and $M$ will be evaluated in our experimental settings in Section~\ref{subsec:exp_components}.

\textbf{Comparing methods.} 
Based on ViT model, we compare our proposed system with various frameworks that employ coreset selection and buffer update strategies. Evaluation is based on Average Accuracy and Average Forgetting. The baseline method, denoted as \textit{None}, is CODA-Prompt~\cite{smith2023coda}, which neither uses coreset selection nor buffer updates. To assess coreset selection, we include rule-based methods such as Camel~\cite{li2022camel}, K-center~\cite{sener2017active}, and FreeSel~\cite{xie2024towards}, as well as model-based approaches like GradMatch~\cite{killamsetty2021grad}, Learn-Loss~\cite{yoo2019learning}, and Craig~\cite{mirzasoleiman2020coresets}. For buffer update strategies, we compare against state-of-the-art methods, including ER~\cite{chaudhry2019tiny}, ASER~\cite{shim2021online}, Camel~\cite{li2022camel}, GSS~\cite{aljundi2019gradient}, and SSD~\cite{gu2024summarizing}.

\textbf{Model Architecture.} 
The architecture is based on the ViT-B/16 backbone~\cite{dosovitskiy2020image}, pre-trained on ImageNet1K~\cite{russakovsky2015imagenet}. We integrate CODA-Prompt~\cite{smith2023coda} in layers 1-5, using a fingerprint length of 8 and a pool of 100 fingerprint components.For optimization, we use Adam with a learning rate of 0.001, set $K=1$ (Eq.~\ref{eq:gd}) for one-step gradient descent per timestamp, and process batches of 20 images with resizing and normalization. The same pre-trained ViT-B/16 backbone is used to extract features, losses, or gradients for different data selection methods. In the Learn-Loss framework, the input dimension is set to 768 with an intermediate dimension of 128 for the loss network.

\begin{table*}[t]
\centering
\caption{
Comparison of state-of-the-art coreset selection and buffer update methods on four datasets, where $\lambda=6028$ samples per second represents the arrival rate of the input data stream. For coreset selection without the auxiliary buffer, we compare our \system method with rule-based methods (Camel, K-center, FreeSel) and model-based methods (GradMatch, Learn-Loss, Craig). Parallelly, based on the coreset selection of \system, we compare five state-of-the-art strategies (ER, ASER, Camel, GSS, SSD) for buffer updates. Note that ``\textbf{Runtime}'' measures the overall running time of each method without the constraint of skipping due to data arrival rates and ``\textbf{-}'' indicates method failure due to excessive overall training time.}
\label{tab:dsbu}
\fontsize{19}{20}\selectfont
\resizebox{0.95\textwidth}{!}{%
\begin{tabular}{cl|ccc|ccc|ccc|ccc}
\toprule
\multicolumn{2}{c|}{\multirow{2}{*}{\textbf{Methods}}} & \multicolumn{3}{c|}{\textbf{Clear10}} & \multicolumn{3}{c|}{\textbf{Clear100}} & \multicolumn{3}{c|}{\textbf{CORe50}} & \multicolumn{3}{c}{\textbf{Stream-51}} \\
\multicolumn{2}{l|}{} & Acc ($\uparrow$) & Fgt ($\downarrow$) & Runtime (s) & Acc ($\uparrow$) & Fgt ($\downarrow$) & Runtime (s) & Acc ($\uparrow$) & Fgt ($\downarrow$) & Runtime (s) & Acc ($\uparrow$) & Fgt ($\downarrow$) & Runtime (s) \\
\midrule
\multicolumn{1}{c|}{\multirow{10}{*}{Coreset Selection}} & None & 29.35 & 23.44 & \white{0}387.75 & \white{0}9.20 & 6.46 & 1291.33 & \white{0}9.33 & \white{0}4.72 & 1407.65 & 33.17 & 13.11 & 1770.73 \\
\multicolumn{1}{c|}{} & Camel & 31.22 & 25.11 & \white{0}326.70 & 11.96 & 8.25 & 1088.01 & 12.58 & 14.25 & 1186.02 & 36.93 & 13.35 & 1491.93 \\
\multicolumn{1}{c|}{} & K-center & 22.26 & 21.44 & \white{0}415.80 & \white{0}8.39 & 6.19 & 1384.74 & \white{0}7.63 & \white{0}9.50 & 1509.48 & 28.19 & 12.71 & 1898.82 \\
\multicolumn{1}{c|}{} & FreeSel & 27.16 & 23.93 & \white{0}407.55 & \white{0}8.35 & 6.22 & 1357.27 & \white{0}9.93 & \white{0}4.38 & 1479.53 & 28.65 & 13.11 & 1861.15 \\
\multicolumn{1}{c|}{} & Learn-Loss & 22.48 & 22.36 & \white{0}412.50 & \white{0}8.48 & 6.37 & 1373.75 & \white{0}7.50 & 10.24 & 1497.50 & 30.40 & 12.82 & 1883.75 \\
\multicolumn{1}{c|}{} & GradMatch & 19.27 & \textbf{17.61} & \white{0}592.35 & \white{0}6.70 & 4.86 & 1972.71 & \white{0}6.74 & \white{0}7.57 & 2150.41 & 16.43 & 12.85 & 2705.07 \\
\multicolumn{1}{c|}{} & Craig & 17.29 & 17.82 & 1465.20 & \white{0}1.76 & \textbf{3.02} & 4879.56 & \white{0}4.03 & \textbf{\white{0}4.03} & 5319.12 & \white{0}3.69 & \textbf{\white{0}9.97} & 6691.08 \\
\rowcolor{lightgray}
\multicolumn{1}{c|}{} & \makecell[l]{\system\\ {\textbf{(w/o buffer)}}}  & \textbf{40.11} & 23.17 & \textbf{\textcolor{lightgray}{0}229.35} & \textbf{19.85} & 8.30 & \textbf{\textcolor{lightgray}{0}763.81} & \textbf{17.70} & 10.99 & \textbf{\textcolor{lightgray}{0}832.61} & \textbf{56.99} & 12.00 & \textbf{1047.37} \\
\midrule
\multicolumn{1}{c|}{\multirow{6}{*}{Buffer Update}} & ER & 54.00 & 0.95 & \textbf{\white{0}427.35} & 35.97 & 4.42 & \textbf{\white{0}1423.21} & 24.07 & 9.23 & \textbf{\white{0}1551.41} & 61.41 & 6.11 & \textbf{\white{0}1951.57} \\
\multicolumn{1}{c|}{} & ASER & 19.92 & 4.32 & 1320.00 & 11.69 & \textbf{0.16} & \white{0}4396.00 & \white{0}6.98 & 3.22 & \white{0}4792.00 & 26.22 & 1.86 & \white{0}6028.00 \\
\multicolumn{1}{c|}{} & Camel & 19.90 & 4.34 & 1994.85 & \white{0}5.96 & 0.49 & \white{0}6643.46 & \white{0}5.09 & \textbf{0.83} & \white{0}7241.91 & 16.16 & 1.28 & \white{0}9109.82 \\
\multicolumn{1}{c|}{} & GSS & - & - & 4600.20 & - & - & 15320.06 & \white{0}6.35 & 1.23 & 16700.12 & \white{0}6.05 & \textbf{0.93} & 21007.58 \\
\multicolumn{1}{c|}{} & SSD & 16.37 & 6.73 & 1577.40 & \white{0}6.04 & 0.36 & \white{0}5253.22 & \white{0}6.90 & 4.43 & \white{0}5726.44 & 19.12 & 2.12 & \white{0}7203.46 \\
\rowcolor{lightgray}
\multicolumn{1}{c|}{} & \system & \textbf{54.94} & \textbf{0.82} & \textcolor{lightgray}{0}448.80 & \textbf{36.57} & 3.02 & \textcolor{lightgray}{0}1494.64 & \textbf{25.24} & 7.68 & \textcolor{lightgray}{0}1629.28 & \textbf{65.78} & 1.53 & \textcolor{lightgray}{0}2049.52 \\
\bottomrule
\end{tabular}%
}
\end{table*}

\subsection{Observation Summary}
We summarize our experimental findings based on a number of observations in the experiment as follows:

\begin{myenumerate}
    \item [F1]
    \textbf{Overall Superior Performance and Adaptability (E2-E5):}
     \system consistently outperforms existing state-of-the-art methods across various experimental settings, demonstrating fast training speed, superior accuracy, and lower forgetting values. With the relatively short runtime, \system achieves accuracy improvements of 3.04\%, 0.33\%, 1.29\%, and 4.45\% respectively over the next best approaches on Clear10, Clear100, CORe50, and Stream-51 datasets (E2). Across different data arrival rates ($\lambda$=30140, 15070, 6028 in E3), \system maintains the highest accuracies of 15.99\%, 33.41\%, and 64.44\% respectively. It also shows robustness to varying class orders (E4), achieving a mean accuracy of 62.44\%, significantly outperforming FreeSel (49.44\%). The consistent superior performance highlights \system's potential for real-world applications in dynamic data scenarios.
    
    \item [F2] \textbf{Indivisual Component Effectiveness (E1 and E6):}
    Component-wise comparison with state-of-the-art methods (E1) demonstrates the superiority of \system's coreset selection and buffer update. With short runtime, our coreset selection improves accuracy by up to 2.95\%, while our buffer update reduces forgetting by up to 4.58\% across datasets. Each component—Fingerprint Attunement (FA), Fingerprint-based Coreset Selection (FCS), and Fingerprint-based Buffer Update (FBU)—significantly enhances performance (E6). The full \system (FA+FCS+FBU) achieves 65.78\% accuracy with only 1.53\% forgetting, showcasing its ability to balance new knowledge acquisition with existing information retention.
    
    \item [F3] \textbf{Optimized Hyperparameter Settings (E7):}
    Optimal performance is achieved with smaller batch sizes (20), enabling finer-grained coreset selection. Lower selection ratios (0.2-0.4), moderate buffer size (204), and fewer gradient descent steps per timestamp (5)  yield high accuracy and reduced forgetting. These results highlight the crucial interplay between computational efficiency, update frequency, and knowledge retention in stream learning.
\end{myenumerate}

\subsection{Experiment Results}
In the following, we present detailed experimental results.

\textbf{E1: Comparison of Coreset Selection and Buffer Update Methods.}
Table~\ref{tab:dsbu} shows \system outperforms conventional coreset selection and buffer update methods across datasets. By using fingerprints as model states, \system selects more informative samples for both coreset and buffer, demonstrating the effectiveness of fingerprint-based learning in dynamic environments.

For coreset selection (upper part of Tab.~\ref{tab:dsbu}), \system substantially outperforms all baselines, demonstrating accuracy improvements of 8.89\%, 7.89\%, 5.12\%, and 20.06\% over the next best methods across Clear10, Clear100, CORe50, and Stream-51 datasets respectively. While \textit{Camel} achieves competitive performance among baselines through its submodular maximization objective function, it still falls short of \system's performance due to its inability to leverage evolving model knowledge. Model-based approaches (\textit{GradMatch}, \textit{Learn-Loss}, \textit{Craig}) show inferior performance, primarily due to their substantial computational overhead and limited adaptability to distribution shifts. Notably, \system achieves these significant improvements while maintaining the lowest computational runtime (229.35s, 763.81s, 832.61s, and 1047.37s), highlighting its efficiency in constructing representative coresets through fingerprint-based selection.

The buffer update comparisons (lower section of Tab.~\ref{tab:dsbu}) demonstrate the superiority of \system across all evaluated datasets, with forgetting reductions of 0.13\%, 1.40\%, 1.55\%, and 4.58\% over the strongest baselines. This consistent improvement can be attributed to our fingerprint-guided sampling mechanism that effectively captures the model's knowledge evolution. While rule-based methods generally exhibit better performance than model-based approaches due to their higher processing capabilities, they remain suboptimal compared to \system. Notably, \textit{ER}, despite its strong performance via reservoir sampling (54.00\%-61.41\%), is constrained by its inherent randomness and inability to leverage model states. Contemporary methods like \textit{GSS} and \textit{SSD} face significant computational challenges, with runtimes ranging from 4600.20s to 21007.58s and occasional failures. In contrast, \system achieves superior accuracy gains of 0.94\%, 0.60\%, 1.17\%, and 4.37\% over \textit{ER} while maintaining comparable computational efficiency. These results validate that our fingerprint-based strategy effectively balances efficiency and performance, facilitating robust knowledge preservation in streaming scenarios.

\begin{table*}[t]
\caption{Comprehensive Comparison of SL Methods. Note that: the \textbf{*} notation indicates that \system serves as either coreset selection for buffer update methods (\textit{ASER}, \textit{GSS}, and \textit{SSD}) or as buffer update for coreset selection methods (\textit{K-center}, \textit{FreeSel}, \textit{Learn-Loss}, \textit{GradMatch}, and \textit{Craig}), since these methods involve only a single component. \textit{ER$^*$} employs random sampling for coreset selection.}
\label{tab:comprehensive}

\begin{subtable}{\textwidth}  
\begin{minipage}{0.62\textwidth}
\centering
\caption{Comparing on four datasets with $\lambda$=6028.}
\label{tab:four_data}
\fontsize{30}{38}\selectfont
\resizebox{1\textwidth}{!}{%
\begin{tabular}{l|ccc|ccc|ccc|ccc}
\toprule
\multirow{2}{*}{\textbf{Method}} &
\multicolumn{3}{c|}{\textbf{Clear10}} &
\multicolumn{3}{c|}{\textbf{Clear100}} &
\multicolumn{3}{c|}{\textbf{CORe50}} &
\multicolumn{3}{c}{\textbf{Stream-51}} \\
& Acc & Fgt & Runtime & Acc & Fgt & Runtime & Acc & Fgt & Runtime & Acc & Fgt & Runtime \\
\midrule
None & 25.21 & 13.68 & \textbf{\white{0}384.45} & \white{0}7.95 & 7.11 & \textbf{\white{0}1280.34} & \white{0}8.10 & 13.02 & \textbf{\white{0}1395.67} & 38.04 & 12.12 & \textbf{\white{0}1755.66} \\
Camel & 22.71 & \white{0}1.39 & 2090.55 & \white{0}5.96 & 0.48 & \white{0}6962.17 & \white{0}5.15 & \textbf{\white{0}0.93} & \white{0}7589.33 & 13.20 & \white{0}0.85 & \white{0}9546.85\\
K-center$^*$ & 50.45 & \white{0}0.50 & \white{0}638.55 & 23.27 & 1.23 & \white{0}2126.57 & 19.81 & \white{0}7.61 & \white{0}2318.13 & 44.90 & \white{0}1.23 & \white{0}2916.05 \\
FreeSel$^*$ & 46.37 & \white{0}0.23 & \white{0}618.75 & 23.27 & 1.70 & \white{0}2060.63 & 20.41 & \white{0}7.46 & \white{0}2246.25 & 46.21 & \white{0}2.10 & \white{0}2825.63 \\
Learn-Loss$^*$ & 49.73 & \white{0}0.62 & \white{0}628.65 & 22.28 & 0.53 & \white{0}2093.60 & 17.67 & \white{0}7.72 & \white{0}2282.19 & 40.72 & \white{0}3.86 & \white{0}2870.84 \\
GradMatch$^*$ & 48.31 & \textbf{\white{0}0.32} & \white{0}815.10 & 16.88 & 0.60 & \white{0}2714.53 & 12.48 & \white{0}6.00 & \white{0}2959.06 & 42.74 & \white{0}1.35 & \white{0}3722.29\\
Craig$^*$ & 22.99 & \white{0}1.38 & 1686.30 & 12.06 & 0.39 & \white{0}5615.89 & \white{0}8.89 & \white{0}4.26 & \white{0}6121.78 & 22.87 & \textbf{\white{0}0.00} & \white{0}7700.77\\
ER$^*$ & 51.90 & \white{0}1.09 & \white{0}412.50 & 36.24 & 4.04 & \white{0}1373.75 & 23.95 & \white{0}9.71 & \white{0}1497.50 & 59.99 & \white{0}3.70 & \white{0}1883.75 \\
ASER$^*$ & 19.92 & \white{0}4.32 & 1320.00 & 11.69 & \textbf{0.16} & \white{0}4396.00 & \white{0}6.98 & \white{0}3.22 & \white{0}4792.00 & 27.82 & \white{0}0.74 & \white{0}6028.00\\
GSS$^*$ & - & - & 4600.20 & - & - & 15320.06 & \white{0}6.35 & \white{0}1.23 & 16700.12 & \white{0}5.99 & \white{0}0.27 & 21007.58 \\
SSD$^*$ & 16.37 & \white{0}6.73 & 1577.40 & \white{0}6.04 & 0.36 & \white{0}5253.22 & \white{0}6.90 & \white{0}4.43 & \white{0}5726.44 & 22.80 & \white{0}1.15 & \white{0}7203.46\\
\rowcolor{lightgray} 
\system & \textbf{54.94} & \textcolor{lightgray}{0}0.82 & \textcolor{lightgray}{0}448.80 & \textbf{36.57} & 3.02 & \textcolor{lightgray}{0}1494.64 & \textbf{25.24} & \textcolor{lightgray}{0}7.68 & \textcolor{lightgray}{0}1629.28 & \textbf{64.44} & \textcolor{lightgray}{0}2.25 & \textcolor{lightgray}{0}2049.52\\
\bottomrule
\end{tabular}%
}
\end{minipage}
\hfill
\begin{minipage}{0.36\textwidth}
\centering
\caption{Comparison on Stream-51 with diverse $\lambda$ values.} 
\label{tab:overall}
\fontsize{15}{17}\selectfont
\resizebox{\textwidth}{!}{%
\begin{tabular}{l|cc|cc|cc}
\toprule
\multirow{2}{*}{\textbf{Method}} & 
\multicolumn{2}{c|}{\textbf{$\lambda$=30140}} &
\multicolumn{2}{c|}{\textbf{$\lambda$=15070}} &
\multicolumn{2}{c}{\textbf{$\lambda$=6028}} \\
& Acc & Fgt & Acc & Fgt & Acc & Fgt \\
\midrule
None & \white{0}4.70 & 5.43 & 11.59 & 11.07 & 38.04 & 12.12 \\
Camel & - & - & \white{0}3.76 & \white{0}2.86 & 13.20 & \white{0}0.85 \\
K-center$^*$ & \white{0}9.57 & 0.47 & 18.40 & \white{0}2.10 & 44.90 & \white{0}1.23 \\
FreeSel$^*$ & \white{0}9.32 & 0.69 & 18.18 & \white{0}0.82 & 46.21 & \white{0}2.10 \\
Learn-Loss$^*$ & \white{0}9.97 & 0.46 & 18.67 & \white{0}1.34 & 40.72 & \white{0}3.86 \\
GradMatch$^*$ & \white{0}6.79 & 0.32 & 13.13 & \textbf{\white{0}0.34} & 42.74 & \white{0}1.35 \\
Craig$^*$ & \white{0}6.07 & \textbf{0.23} & \white{0}4.96 & \white{0}2.23 & 22.87 & \white{0}0.00 \\
ER$^*$ & 14.18 & 0.77 & 30.89 & \white{0}1.03 & 59.99 & \white{0}3.70 \\
ASER$^*$ & \white{0}5.99 & 0.27 & \white{0}8.52 & \white{0}1.63 & 27.82 & \white{0}0.74 \\
GSS$^*$ & - & - & - & - & \white{0}5.99 & \textbf{\white{0}0.27} \\
SSD$^*$ & \white{0}5.37 & 0.40 & \white{0}5.88 & \white{0}2.84 & 22.80 & \white{0}1.15 \\
\rowcolor{lightgray}
\system & \textbf{15.99} & 0.36 & \textbf{33.41} & \textcolor{lightgray}{0}0.68 & \textbf{64.44} & \textcolor{lightgray}{0}2.25 \\
\bottomrule
\end{tabular}}
\end{minipage}
\end{subtable}

\vspace{3pt}

\begin{subtable}{\textwidth}
\centering
\caption{Comparison under different class orders on Stream-51 with $\lambda$=6028.}
\label{tab:cls_ord}
\fontsize{8}{9}\selectfont
\resizebox{0.87\textwidth}{!}{%
\begin{tabular}{l|cc|cc|cc|cc|cc|cc}
\toprule
\multirow{2}{*}{\textbf{Method}} & \multicolumn{2}{c|}{\textbf{Class order 1}} & \multicolumn{2}{c|}{\textbf{Class order 2}} & \multicolumn{2}{c|}{\textbf{Class order 3}} & \multicolumn{2}{c|}{\textbf{Class order 4}} & \multicolumn{2}{c|}{\textbf{Class order 5}} & \multicolumn{2}{c}{\textbf{Overall}} \\
 & Acc & Fgt & Acc & Fgt & Acc & Fgt & Acc & Fgt & Acc & Fgt & Acc & Fgt \\
\midrule
None & 38.04 & 12.12 & 34.72 & 10.16 & 34.70 & 10.93 & 27.88 & 10.20 & 32.87 & 11.99 & 33.64\scriptsize{$\pm$4.62} & 11.08\scriptsize{$\pm$1.17} \\
Camel & 13.20 & \white{0}0.85 & 15.51 & \white{0}1.43 & 11.31 & \white{0}1.66 & 17.25 & \white{0}2.13 & 14.73 & \white{0}2.01 & 14.40\scriptsize{$\pm$2.81} & \white{0}1.62\scriptsize{$\pm$0.63} \\
K-center$^*$ & 44.90 & \white{0}1.23 & 53.64 & \white{0}2.53 & 49.33 & \white{0}2.15 & 46.57 & \white{0}6.46 & 48.26 & \white{0}2.30 & 48.54\scriptsize{$\pm$4.11} & \white{0}2.93\scriptsize{$\pm$2.52} \\
FreeSel$^*$ & 46.21 & \white{0}2.10 & 51.30 & \white{0}2.11 & 49.91 & \white{0}3.67 & 48.77 & \white{0}6.26 & 50.99 & \white{0}0.41 & 49.44\scriptsize{$\pm$2.56} & \white{0}2.91\scriptsize{$\pm$2.73} \\
Learn-Loss$^*$ & 40.72 & \white{0}3.86 & 54.48 & \white{0}2.23 & 50.09 & \white{0}3.35 & 45.96 & \white{0}7.11 & 51.85 & \white{0}1.36 & 48.62\scriptsize{$\pm$6.70} & \white{0}3.58\scriptsize{$\pm$2.73} \\
GradMatch$^*$ & 42.74 & \white{0}1.35 & 41.85 & \white{0}2.13 & 40.96 & \white{0}2.20 & 40.79 & \white{0}3.53 & 40.29 & \white{0}1.32 & 41.33\scriptsize{$\pm$1.21} & \white{0}2.11\scriptsize{$\pm$1.12} \\
Craig$^*$ & 22.87 & \textbf{\white{0}0.00} & 17.80 & \white{0}1.55 & 19.08 & \white{0}2.26 & 18.95 & \textbf{\white{0}0.93} & 19.19 & \textbf{\white{0}0.41} & 19.58\scriptsize{$\pm$2.39} & \textbf{\white{0}1.03\scriptsize{$\pm$1.12}} \\
ER$^*$ & 59.99 & \white{0}3.70 & 60.60 & \white{0}2.27 & 57.57 & \white{0}6.49 & 54.89 & \white{0}9.19 & 58.11 & \white{0}6.76 & 58.23\scriptsize{$\pm$2.80} & \white{0}5.68\scriptsize{$\pm$3.38} \\
ASER$^*$ & 27.82 & \white{0}0.74 & 24.07 & \textbf{\white{0}0.91} & 25.68 & \textbf{\white{0}0.59} & 29.01 & \white{0}3.26 & 23.81 & \white{0}1.04 & 26.08\scriptsize{$\pm$2.84} & \white{0}1.31\scriptsize{$\pm$1.37} \\
GSS$^*$ & \white{0}5.99 & \white{0}0.27 & \white{0}7.11 & \white{0}1.69 & \white{0}4.69 & \white{0}1.63 & \white{0}5.92 & \white{0}1.77 & \white{0}6.72 & \white{0}1.39 & \white{0}6.09\scriptsize{$\pm$1.15} & \white{0}1.35\scriptsize{$\pm$0.77} \\
SSD$^*$ & 22.80 & \white{0}1.15 & 20.35 & \white{0}2.35 & 22.10 & \white{0}1.38 & 16.22 & \white{0}1.95 & 24.11 & \white{0}0.85 & 21.12\scriptsize{$\pm$3.79} & \white{0}1.54\scriptsize{$\pm$0.75} \\
\rowcolor{lightgray}
\system & \textbf{64.44} & \textcolor{lightgray}{0}2.25 & \textbf{62.31} & \textcolor{lightgray}{0}2.56 & \textbf{62.20} & \textcolor{lightgray}{0}3.74 & \textbf{62.81} & \textcolor{lightgray}{0}6.61 & \textbf{60.44} & \textcolor{lightgray}{0}1.85 & \textbf{62.44\scriptsize{$\pm$1.78}} & \textcolor{lightgray}{0}3.40\scriptsize{$\pm$2.39} \\
\bottomrule
\end{tabular}%
}
\end{subtable}
\end{table*}

\textbf{E2: Comparison of SL Methods on Four Datasets.} 
Table~\ref{tab:four_data} comprehensively evaluates \system against state-of-the-art SL methods across four datasets at $\lambda$=6028, demonstrating our method's advantages in both effectiveness and efficiency:

In terms of effectiveness, \system achieves superior accuracy (54.94\%, 36.57\%, 25.24\%, and 64.44\% on Clear10, Clear100, CORe50, and Stream-51), consistently outperforming the strongest baseline \textit{ER$^*$} by margins of 3.04\%, 0.33\%, 1.29\%, and 4.45\% respectively. More importantly, \system exhibits substantially lower forgetting rates across all datasets: achieving 0.82\% on Clear10 compared to \textit{ER$^*$} (1.09\%) and \textit{Camel} (1.39\%), and 2.25\% on Stream-51 versus \textit{ER$^*$} (3.70\%) and \textit{Learn-Loss$^*$} (3.86\%). In contrast, existing methods show significant performance limitations: model-based approaches like \textit{GradMatch$^*$} and \textit{Craig$^*$} suffer substantial accuracy degradation on challenging datasets (achieving only 12.48\% and 8.89\% on CORe50), while \textit{ASER$^*$} consistently underperforms across all datasets (19.92\%, 11.69\%, 6.98\%, and 27.82\%).

Regarding computational efficiency, \system maintains stable runtimes (448.80s, 1494.64s, 1629.28s, and 2049.52s) across all datasets, comparable to simple baselines like \textit{None} and \textit{ER$^*$}. In contrast, some methods face severe computational challenges: \textit{GSS$^*$} fails to complete on Clear10 and Clear100 due to excessive runtime (reaching 21007.58s on Stream-51), and \textit{SSD$^*$} suffers from high computational overhead (1577.40s-7203.46s). 
These comprehensive results validate that our synergistic design, which combines adaptive coreset selection for balanced knowledge acquisition and dynamic buffer updates for optimized memory utilization, enables robust model learning with both superior performance and computational efficiency, demonstrating strong practical viability for real-world streaming scenarios.


\textbf{E3: Comparison of SL Methods on Data arrival rates.} 
Table~\ref{tab:overall} evaluates \system against state-of-the-art SL methods on Stream-51 under varying data arrival rates ($\lambda$), demonstrating our method's robustness and superiority. In terms of accuracy, \system consistently outperforms existing methods across all arrival rates. At high arrival rate ($\lambda$=30140), \system achieves 15.99\% accuracy while maintaining a low forgetting rate of 0.36\%, surpassing \textit{ER$^*$} (14.18\% accuracy, 0.77\% forgetting). Notably, several methods including \textit{Camel}, \textit{GSS$^*$} and \textit{ASER$^*$} fail to handle such rapid data streams. At medium arrival rate ($\lambda$=15070), \system reaches 33.41\% accuracy with 0.68\% forgetting, significantly outperforming the accuracy of  \textit{ER$^*$} (30.89\%) and \textit{Learn-Loss$^*$} (18.67\%). The advantage becomes more pronounced at lower arrival rate ($\lambda$=6028), where \system attains 64.44\% accuracy, substantially exceeding \textit{ER$^*$} (59.99\%) and \textit{FreeSel$^*$} (46.21\%).

The performance trend across arrival rates reveals an interesting pattern: while all methods benefit from lower arrival rates (more processing time per batch), \system maintains the largest performance margins (improvements of 1.81\%, 2.52\%, and 4.45\% over \textit{ER$^*$}) and the stable forgetting values (0.36\%, 0.68\%, and 2.25\%). This robust scaling can be attributed to our synergistic design, which combines adaptive coreset selection for balanced knowledge acquisition and dynamic buffer updates for optimized memory utilization, enabling effective continual learning even under challenging streaming conditions.

\begin{table*}[t]
\centering
\caption{Comparison of the settings with and without skipping batches under five class orders on Stream-51 with $\lambda$=300.}
\label{tab:skip}
\fontsize{7}{8}\selectfont
\resizebox{0.87\textwidth}{!}{%
\begin{tabular}{l|cc|cc|cc|cc|cc|cc}
\toprule
\multirow{2}{*}{\textbf{Setting}} & \multicolumn{2}{c|}{\textbf{Class order 1}} & \multicolumn{2}{c|}{\textbf{Class order 2}} & \multicolumn{2}{c|}{\textbf{Class order 3}} & \multicolumn{2}{c|}{\textbf{Class order 4}} & \multicolumn{2}{c|}{\textbf{Class order 5}} & \multicolumn{2}{c}{\textbf{Overall}} \\
 & Acc & Fgt & Acc & Fgt & Acc & Fgt & Acc & Fgt & Acc & Fgt & Acc & Fgt \\
 \midrule
w/o skipping & 64.11 & 31.03 & 64.91 & 36.51 & 63.76 & 37.61 & 59.50 & 44.86 & 71.25 & 29.00 & 64.71\scriptsize{$\pm$5.24} & 35.80\scriptsize{$\pm$7.73} \\
w/ skipping & \textbf{73.04} & \textbf{30.28} & \textbf{74.45} & \textbf{27.89} & \textbf{77.11} & \textbf{27.10} & \textbf{68.54} & \textbf{37.19} & \textbf{77.16} & \textbf{25.03} & \textbf{74.06\scriptsize{$\pm$4.42}} & \textbf{29.50\scriptsize{$\pm$5.83}}\\
\bottomrule
\end{tabular}%
}
\end{table*}

\textbf{E4: Comparison of Class Orders.}
In streaming scenarios, the sequential order of class appearances significantly impacts model performance. Performance variations across different class orders stem from the varying degrees of knowledge transferability between consecutive tasks. When consecutive tasks exhibit higher similarity, knowledge transfer tends to be more effective; conversely, dissimilar tasks may impede knowledge transfer and degrade performance.

Experimental results in Table~\ref{tab:cls_ord} validate the effectiveness of \system across diverse class orders. \system consistently outperforms baseline methods, achieving 62.44\% mean accuracy compared to 58.43\% (\textit{ER$^*$}) and 49.44\% (\textit{FreeSel$^*$}). More importantly, \system exhibits superior stability with only 4\% accuracy variation (60.44\%-64.44\%) across different orders, while \textit{ER$^*$} shows larger fluctuations (54.89\%-60.60\%). In terms of catastrophic forgetting, \system achieves a mean forgetting rate of 3.40\%, substantially outperforming \textit{ER$^*$} (5.68\%). The effectiveness of \system is further highlighted when compared to \textit{None} method without any strategy (33.64\% accuracy, 11.08\% forgetting).

The superior performance of \system can be attributed to its effective balance between knowledge acquisition and retention, achieved through strategically designed coreset selection and buffer update mechanisms. This is evidenced by its robust performance in challenging scenarios, particularly in Class order 4 where \system maintains 62.81\% accuracy while other methods deteriorate significantly. Unlike methods such as \textit{CSS$^*$} and \textit{ASER$^*$} that achieve low forgetting rates at the cost of accuracy, \system effectively addresses the stability-plasticity trade-off in streaming scenarios.

\textbf{E5: Comparison of Different Skipping Setting.}
Following prior work~\cite{ghunaim2023real}, we investigated two data processing strategies in streaming scenarios. The first strategy employs batch skipping: when the model cannot process incoming data in real-time, entire batches are skipped while maintaining a high coreset selection ratio for the processed batches. The second strategy processes all batches but with a lower sample selection ratio per batch, ensuring real-time processing capability while preserving the continuity of the data stream.

The results demonstrate that the batch-skipping configuration consistently yields superior performance. The average accuracy improves significantly from 64.71\% to 74.06\% (+9.35\%), while the forgetting metric decreases from 35.80\% to 29.50\% (-6.30\%). This improvement is consistent across all class orders, with the most substantial accuracy gains observed in Class order 3 (from 63.76\% to 77.11\%, +13.35\%) and Class order 4 (from 59.50\% to 68.54\%, +9.04\%).
These results suggest that, although batch skipping is not ideal, it allows the model to prioritize more relevant data, thereby enhancing training efficiency and performance. While skipping shows better results, a fixed selection ratio may not be optimal. We propose exploring a dynamic selection ratio across batches as a potential direction for future work.


\subsection{Effectiveness of Key Components}
\label{subsec:exp_components}
\textbf{E6: Ablation Studies.} We present a detailed ablation study in Table~\ref{tab:aba} using \system with a 0.5 coreset selection ratio and a buffer size of 102 on Stream-51 with $\lambda=6028$. The table demonstrates that each component of \system contributes to improved performance. The first row, where no components are applied, serves as the baseline, showing an accuracy of 33.29\% and a forgetting of 15.41\%. Adding fingerprint attunement (FA) alone, as seen in the second row, increases accuracy to 35.69\% and reduces forgetting to 13.82\% with a slight time cost of 37.68s, indicating that FA enhances model inference by improving the quality of fingerprints. In the third row, both fingerprint attunement and fingerprint-based coreset selection (FCS) are employed, leading to a substantial improvement in accuracy to 56.74\% and a reduction in forgetting to 11.35\%. This shows that FCS significantly boosts model performance by selecting more informative data and speeding up the training process, allowing the model to make better use of the fast-stream data. The final row illustrates the combined effect of all components, including FA, FCS, and FBU (fingerprint-based buffer update). Here, the accuracy peaks at 65.78\% and the forgetting significantly drops to 1.53\%, suggesting FBU is highly effective in both acquiring new knowledge and retaining previously learned information. Overall, compared with the baseline, with only an additional time of 346.61s, our method improved accuracy by 32.49\% while reducing forgetting by 13.88\%. The ablation study shows all \system components are essential, with their combination yielding optimal performance.

\begin{table}[t]
\centering
\caption{Ablation study on Stream-51 with $\lambda$=6028. \textbf{FA}, \textbf{FCS}, and \textbf{FBU} represent the components of \system: fingerprint attunement, fingerprint-based coreset selection, and fingerprint-based buffer update. The table shows the contribution of each component to accuracy and forgetting.}
\label{tab:aba}
\fontsize{10}{11}\selectfont
\resizebox{0.32\textwidth}{!}{%
\begin{tabular}{ccc|ccc}
\toprule
\multicolumn{1}{c}{\textbf{FA}} & \multicolumn{1}{c}{\textbf{FCS}} & \multicolumn{1}{c|}{\textbf{FBU}} & Acc & Fgt & Runtime \\
\midrule
\xmark & \xmark & \xmark & 33.29 & 15.41 & 1702.91\\
\cmark & \xmark & \xmark & 35.69 & 13.82 & 1740.59\\
\cmark & \cmark & \xmark & 56.74 & 11.35 & 1047.37\\
\cmark & \cmark & \cmark & 65.78 & \white{0}1.53 & 2049.52\\
\bottomrule
\end{tabular}%
}
\end{table}

\begin{figure*}[t]
    \centering
    \begin{subfigure}[b]{0.245\textwidth}
        \centering
        \includegraphics[width=\textwidth]{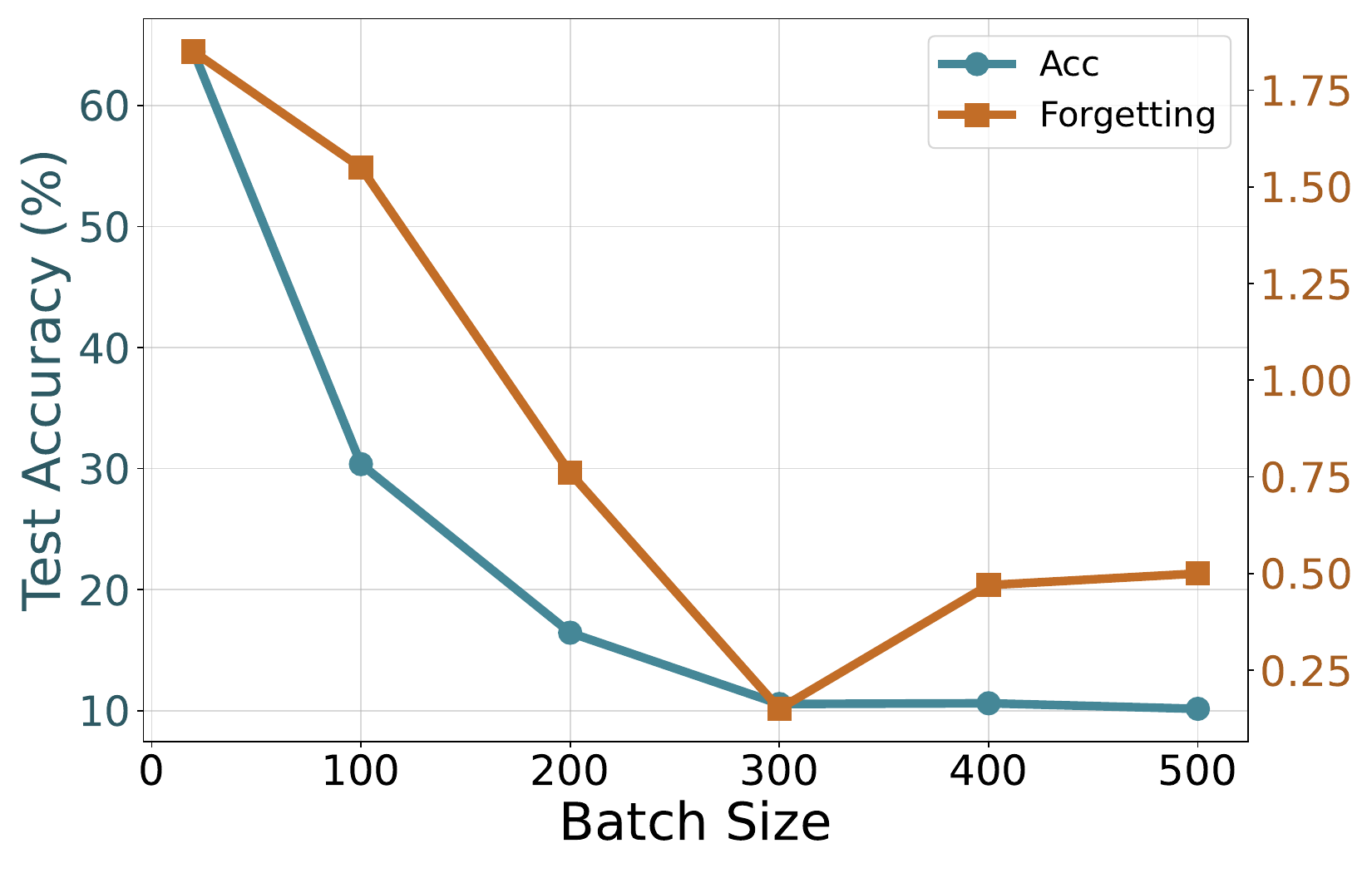}
        \caption{Impact of Batch Sizes $b$}
        \label{fig:sens_bat}
    \end{subfigure}
    \begin{subfigure}[b]{0.245\textwidth}
        \centering
        \includegraphics[width=\textwidth]{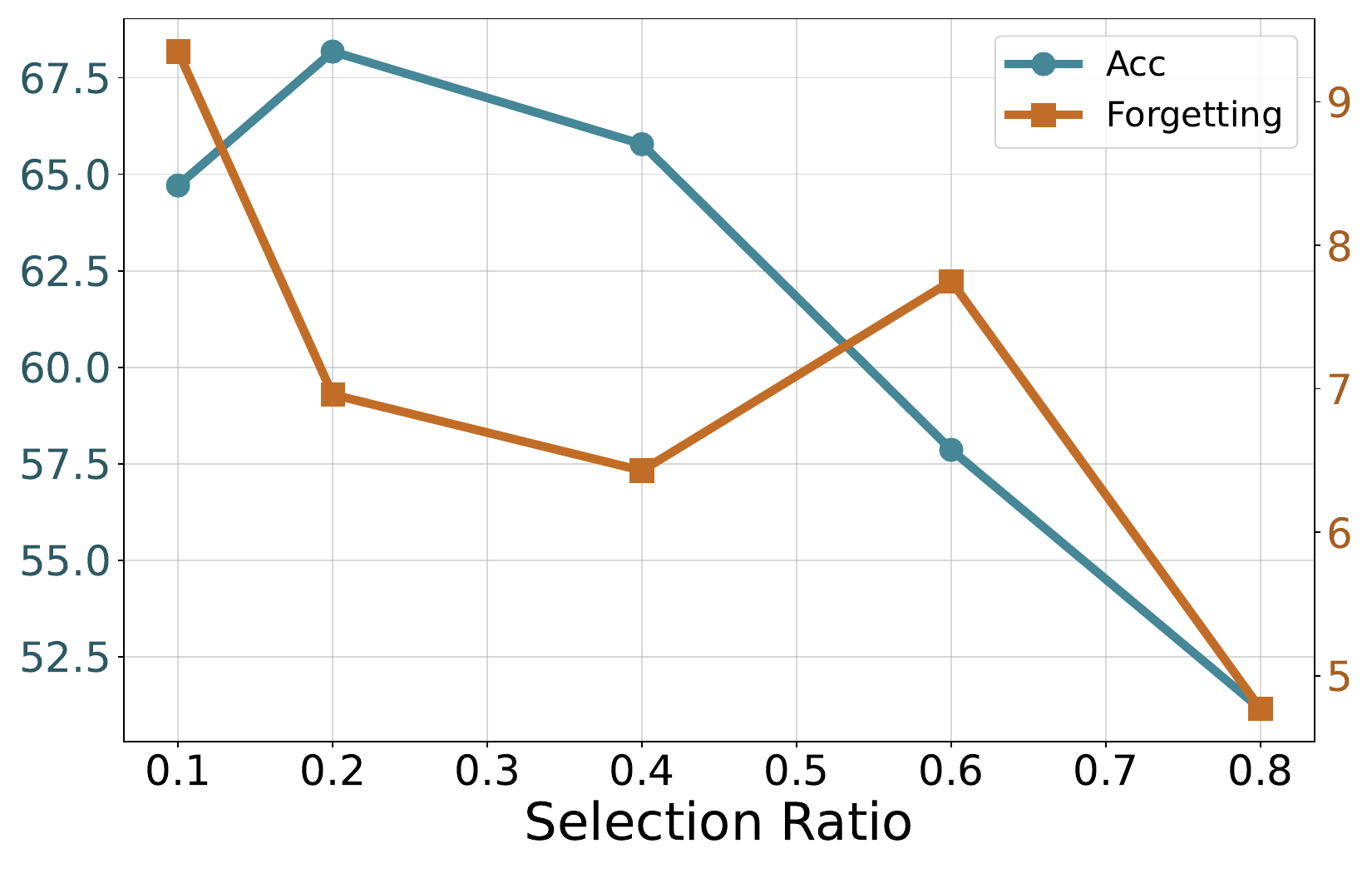}
        \caption{Impact of Selection Ratio $\sigma$}
        \label{fig:sens_sr}
    \end{subfigure}
    \begin{subfigure}[b]{0.245\textwidth}
        \centering
        \includegraphics[width=\textwidth]{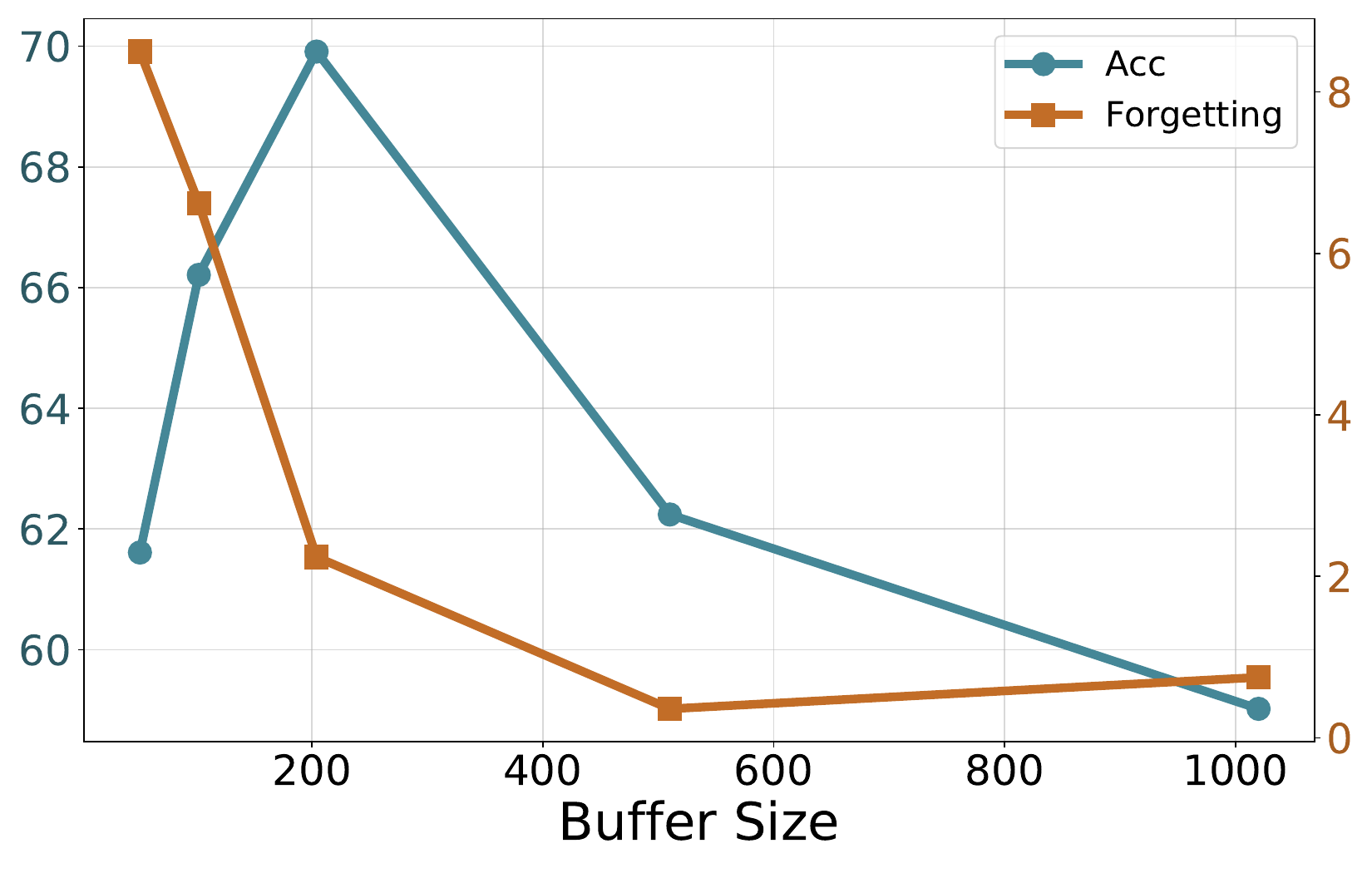}
        \caption{Impact of Buffer Sizes $m$}
        \label{fig:sens_buf}
    \end{subfigure}
    \begin{subfigure}[b]{0.249\textwidth}
        \centering
        \includegraphics[width=\textwidth]{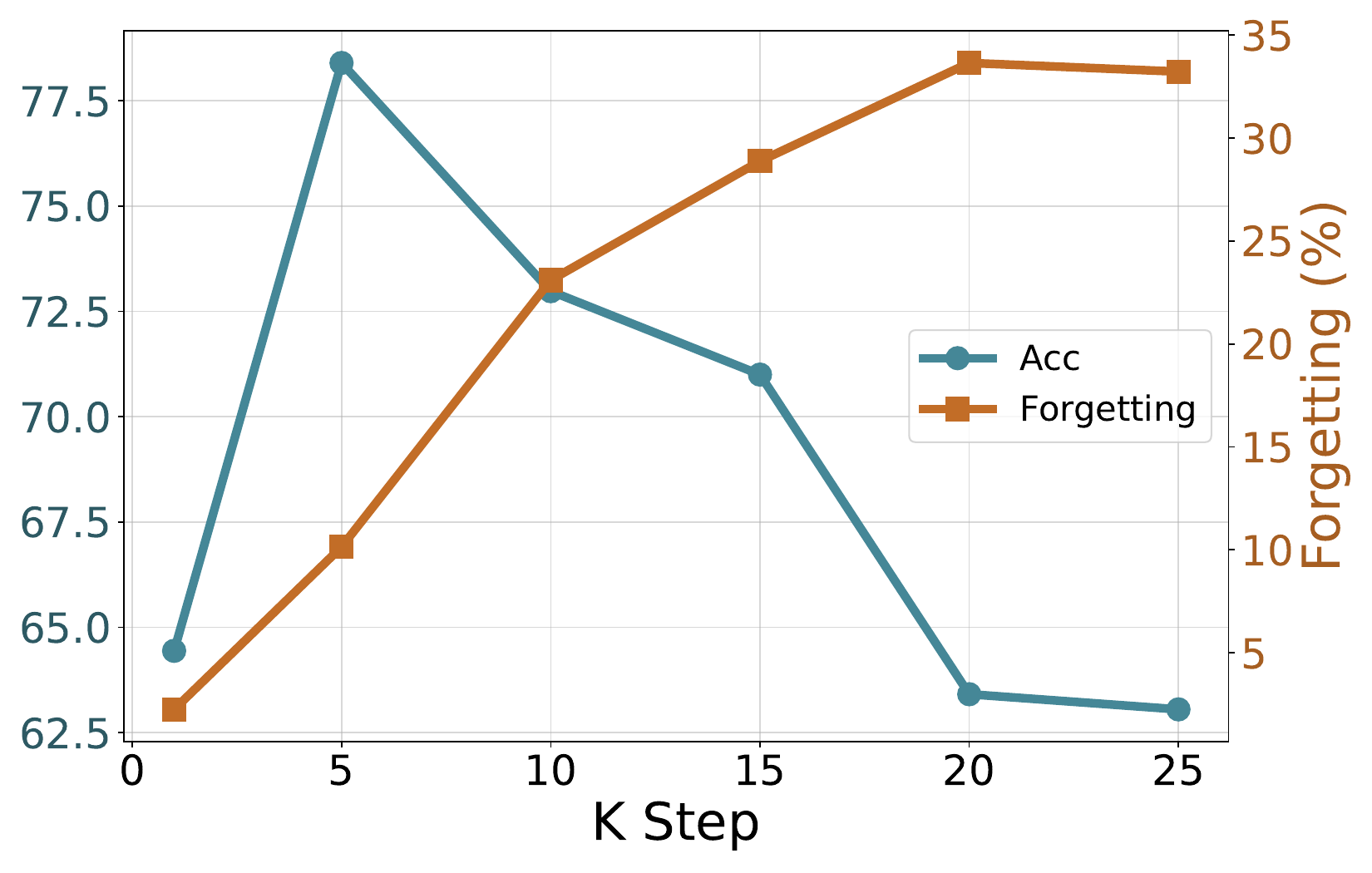}
        \caption{Impact of $K$ steps in Eq.~\ref{eq:gd}.}
        \label{fig:sens_k}
    \end{subfigure}
    \caption{Sensitivity study on Stream-51 with $\lambda$=6028.}
    \label{fig:sensitive}
\end{figure*}

\textbf{E7: Sensitivity Study.} 
We analyze the impact of batch size $b$, coreset selection ratio $\sigma$, buffer size $m$, and gradient descent steps $K$ per timestamp on model accuracy and forgetting, as shown in Figure~\ref{fig:sensitive}.
Fig.~\ref{fig:sens_bat} shows model performance declines as batch size increases from 20 to 500, with optimal results at batch size 20. This is because smaller batches enable more fine-grained coreset selection, allowing better identification of informative samples.
In Fig.~\ref{fig:sens_sr}, lower selection ratios $\sigma$ (0.2 and 0.4) yield high accuracy with reduced forgetting. As $\sigma$ increases to 0.8, accuracy declines while forgetting fluctuates. This result suggests that selecting fewer data batches for model updates during rapid data streams optimizes the balance between performance and knowledge retention.
Fig.~\ref{fig:sens_buf} shows the impact of buffer size $m$, where moderate sizes (102, 204) achieve higher accuracy ($>66\%$) than larger ones (510, 1020). A buffer size of 204 provides the optimal balance between accuracy and forgetting. Larger buffers reduce forgetting but offer diminishing accuracy gains, due to increased maintenance overhead.
Finally, fixing the buffer size ($m=102$), Fig.~\ref{fig:sens_k} shows that both accuracy and forgetting metrics deteriorate with increasing gradient descent steps ($K\geq10$). Optimal performance is achieved at $K=5$, yielding the highest accuracy of 78.40\% and minimal forgetting of 10.14\%. This phenomenon can be attributed to the limited buffer size, which causes the model to overwrite previously learned knowledge when subjected to excessive repeated updates. For sanity checks, we further conducted additional experiments with varying buffer sizes ($m=51$, $m=240$, and $m=510$) and observed that the corresponding optimal number of gradient descent steps $K$ increased to 5, 10, and 15. This observation suggests that larger buffers can accommodate more updates without significantly overwriting prior knowledge, thereby allowing the model to benefit from additional training steps. However, the diminishing returns of larger buffers and increased $K$ highlight the importance of tuning these parameters in concert with the system's computational constraints.

%% file: Body/7_RelatedWork.tex
\section{Related Work} \label{sec:related}
We categorize recent advancements in the field into three areas: \textit{Continual Learning}, \textit{Prompting for Continual Learning}, and \textit{Stream Learning}, each contrasting with our approach.

\textbf{Online Continual Learning (OCL).} Continual Learning (CL) focuses on avoiding catastrophic forgetting, where learning new data overwrites prior knowledge. Approaches include rehearsal methods, which retain or synthesize past data, and regularization techniques that preserve critical parameters to balance flexibility and performance~\cite{lopez2017gradient, kirkpatrick2017overcoming}. Dynamic architectures, such as network expansion or modularization, further mitigate forgetting~\cite{rusu2016progressive, mallya2018packnet}. OCL adapts these strategies for single-pass, sequential learning over shuffled datasets. However, while OCL minimizes reliance on data re-exposure, it doesn't fully address the complexities of high-throughput environments in \textit{stream learning} (SL).

\textbf{Prompting for Continual Learning.} 
The Transformer architecture has emerged as a cornerstone of modern machine learning, demonstrating remarkable success across domains~\cite{brown2020language, dosovitskiy2020image, devlin2018bert}. Prompt tuning methods, which augment this frozen architecture with learnable parameters (prompts), have shown exceptional performance in both NLP~\cite{10.1145/3560815} and CV tasks~\cite{jia2022visual} by enabling efficient adaptation of pre-trained models to fine-tune on unseen data. In continual learning, these methods have evolved significantly, as evidenced by L2P's learnable parameters~\cite{wang2022learning}, DualPrompt's task-specific knowledge separation~\cite{wang2022dualprompt}, CODA-Prompt's attention mechanisms~\cite{smith2023coda}, and HiDe-Prompt's hierarchical learning approach~\cite{wang2024hierarchical}. Despite the prevalence of prompt-based methods, their efficacy in handling high-velocity streaming data remains largely unexplored. While prior work has focused primarily on optimizing these learnable parameters for model effectiveness, our framework uniquely integrates them with data selection strategies, simultaneously enhancing model adaptability and computational efficiency in dynamic data streams.

\textbf{Stream Learning (SL).} SL presents challenges that go beyond CL and OCL due to its unbounded, dynamic data streams and high computational demands~\cite{zhou2024learnabilitytimesharingcomputationalresource}.  Recent advancements include enhanced buffer mechanisms and real-time benchmarks to improve training efficiency~\cite{chaudhry2019tiny, aljundi2019online, he2021online, 10203627, wu-etal-2023-sentistream}. Camel~\cite{li2022camel} exemplifies the push toward better coreset selection by taking the submodular maximization as an object function. Rule-based methods in SL struggle with quick adaptation, while model-based approaches, though more responsive, often carry high computational costs~\cite{sener2017active, mirzasoleiman2020coresets, yoo2019learning, killamsetty2021grad}. Managing the risks of catastrophic forgetting and coping with computational demands in dynamic environments are central challenges~\cite{aljundi2019gradient, aljundi2019online, jin2021gradient}. Based on the widespread ViT architecture, our approach introduces dynamic, learnable fingerprints to SL, which adapts to ever-evolving data streams, enhancing real-time processing and computational efficiency.

%% file: Body/8_Conclusion.tex
\section{Conclusion}
\label{sec:conclusion}
This paper presents a novel framework named \system for efficient stream learning (SL) that achieves both effectiveness and efficiency. \system introduces three innovative strategies: fingerprint attunement to refine fingerprints, selection of coresets based on fingerprints to improve data efficiency, and buffer updates based on fingerprints to optimize training effectiveness, which collectively represent a significant advancement in stream learning methodologies. Extensive experiments on various datasets demonstrate that our method significantly improves existing SL methods in terms of accuracy, adaptivity, and training throughput. \system's dynamic, learnable fingerprints, ensure the model can seamlessly integrate new information while retaining previously learned knowledge, effectively addressing the core challenges of SL. Furthermore, \system's fingerprint attunement leverages the robustness of attention layers in Vision Transformers (ViTs), enhancing the efficiency of fingerprint learning with minimal computational overhead. These innovations collectively establish \system as a scalable and effective solution for the evolving demands of SL, supported by empirical results that demonstrate its performance in varying data sizes and complexities. 
While the current implementation focuses on ViTs, the methodology underlying \system is highly adaptable and can be extended to other transformer architectures with minimal modifications, paving the way for broader applicability in future work.

%% file: Body/Appendix.tex
\section{Appendix}
\subsection{Gradient Correlations between different Tasks} \label{appendix:grad}
\begin{figure}[t]
    \centering
    \begin{subfigure}{\linewidth}
        \centering
        \includegraphics[width=0.75\textwidth]{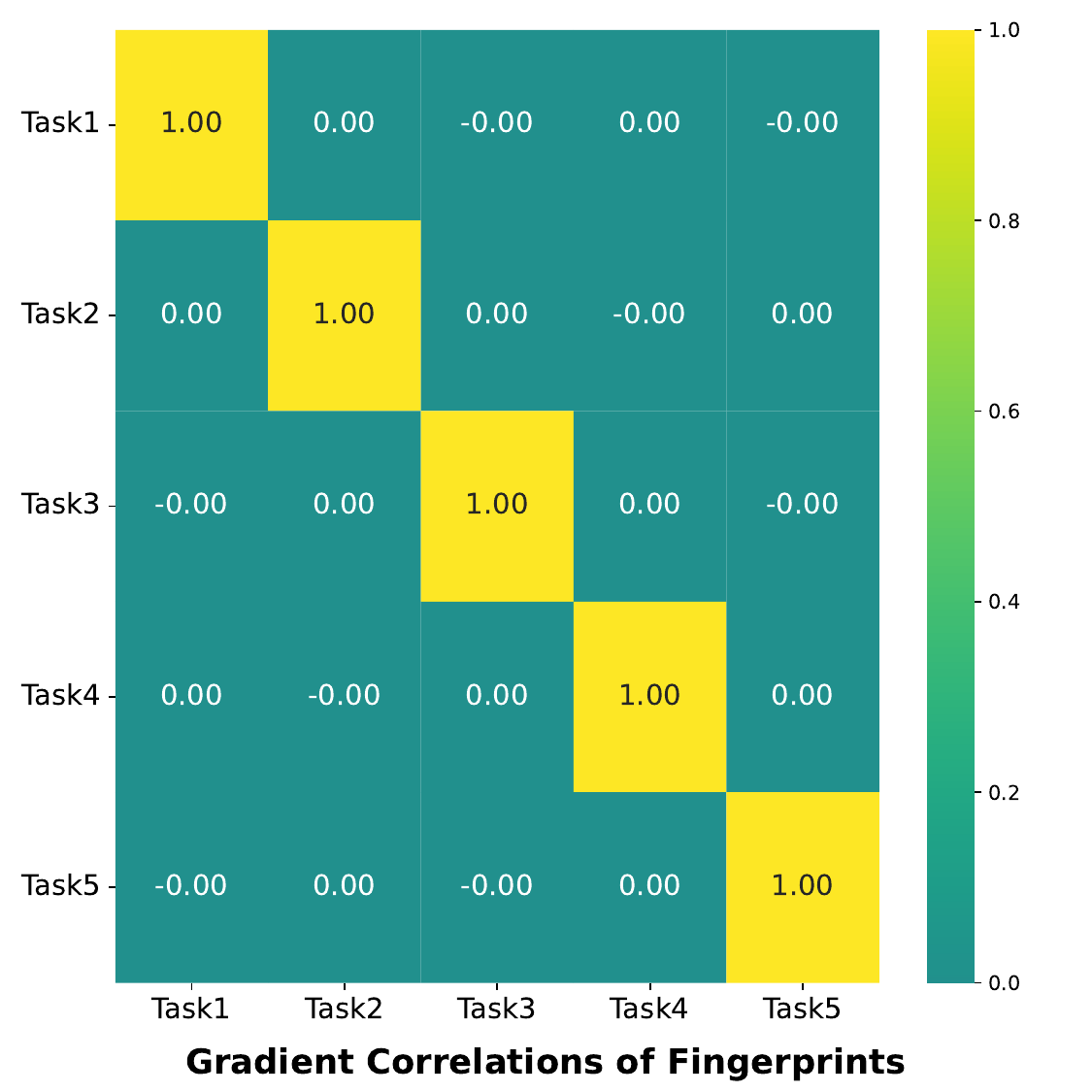}
        \label{fig:grad_prompt}
    \end{subfigure}
    \begin{subfigure}{\linewidth}
        \centering
        \includegraphics[width=0.75\textwidth]{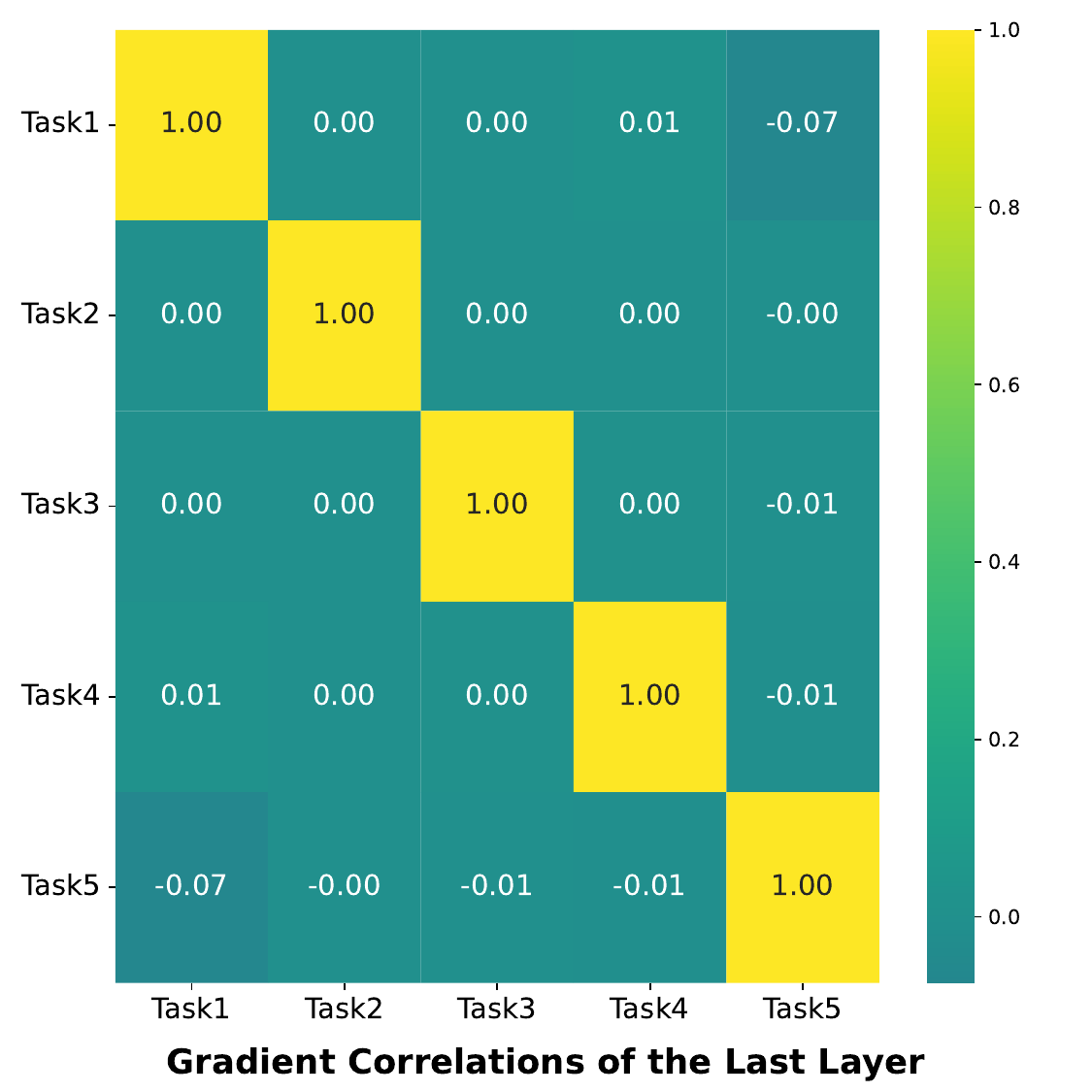}
        \label{fig:grad_lastlayer}
    \end{subfigure}
    \caption{Heatmap of gradient correlations for diverse tasks on Stream-51.}
    \label{fig:grad}
\end{figure}
As demonstrated in Fig.~\ref{fig:grad}, the gradients of both fingerprints and the last layer demonstrate strong orthogonality across distinct tasks, as evidenced by near-zero correlation coefficients in off-diagonal elements. This gradient orthogonality indicates that fingerprint parameters evolve along task-specific trajectories during streaming learning. The gradient correlation analysis of the last layer further reveals that independently optimized fingerprints, leveraging model-specific knowledge, effectively minimize cross-task interference in the final layer outputs.

\subsection{Comparison of Different Pretrained Models.}
\label{appendix:pretrained_models}
\begin{table*}[!t]
\centering
\caption{Comparison of ER$^*$ and \system based on different pretrained models on Stream-51 with $\lambda$=6028.}
\label{tab:ptm}
\fontsize{8}{9}\selectfont
\resizebox{0.8\textwidth}{!}{%
\begin{tabular}{l|cc|cc|cc|cc|cc|cc}
\toprule
\multirow{2}{*}{\textbf{Method}} & \multicolumn{2}{c|}{\textbf{ImageNet-1K}} & \multicolumn{2}{c|}{\textbf{ImageNet-21K}} & \multicolumn{2}{c|}{\textbf{DINO-1K}} & \multicolumn{2}{c|}{\textbf{iBOT-1K}} & \multicolumn{2}{c|}{\textbf{iBOT-21K}} & \multicolumn{2}{c}{\textbf{MoCo-1K}} \\
 & Acc & Fgt & Acc & Fgt & Acc & Fgt & Acc & Fgt & Acc & Fgt & Acc & Fgt \\
\midrule
ER$^*$ & 59.99 & 3.70 & 55.68 & 2.09 & 44.83 & 5.43 & 39.07 & 2.50 & 2.12 & 9.71 & 2.00 & 10.03 \\
\rowcolor{lightgray}
\system & \textbf{64.44} & \textbf{2.25} & \textbf{60.45} & \textbf{0.00} & \textbf{47.44} & \textbf{4.91} & \textbf{41.79} & \textbf{1.35} & \textbf{3.12} & \textbf{8.00} & \textbf{2.52} & \textbf{\textcolor{lightgray}{0}7.01} \\
\bottomrule
\end{tabular}%
}
\end{table*}
As \system is based on the pretrained ViT, we also experiment with varying pretrained models based on the supervised (ImageNet-1K~\cite{russakovsky2015imagenet} and ImageNet-21K~\cite{ridnik2021imagenet}) and the self-supervised (iBOT~\cite{zhou2021ibot}, DINO~\cite{caron2021emerging}, and MoCo v3~\cite{chen2021empirical}) datasets. Results are shown in Table~\ref{tab:ptm}. \system consistently outperforms the \textit{ER$^*$} method across different pretrained models in terms of accuracy and forgetting. Specifically, when using the ImageNet-1K pretrained model, \system achieves the highest accuracy of 64.44\%, compared to \textit{ER$^*$}’s 59.99\%. With ImageNet-21K, \system even achieves the 0\% forgetting that preserves all learned knowledge in the streaming setting. Notably, the performance differences highlight the impact of different pretraining strategies on stream learning outcomes, with \system consistently providing significant improvements in accuracy, albeit with varying degrees of forgetting. These results underscore the robustness and versatility of \system across different pretraining contexts.

\subsection{Proof of Coreset Quality Guarantee} \label{appendix:proof_cqg}
We prove that our fingerprint-based coreset selection method satisfies the standard definition of coreset with rigorous theoretical guarantees.
\begin{theorem}[Coreset Quality Guarantee]
With probability at least $1-\delta$, the coreset $C^t$ satisfies:
\[(1-\varepsilon)\text{cost}(B^t) \leq \text{cost}(C^t) \leq (1+\varepsilon)\text{cost}(B^t),\]
where $\text{cost}(X) = \frac{1}{|X|}\sum_{x\in X} d(x)$, representing the average angular distance, $d(x) = \arccos(\text{sim}(x,P))$, and $\varepsilon = O(\sqrt{\log(1/\delta)/(\sigma b)})$.
\end{theorem}

\begin{proof}
Let us first define the notations:
\begin{itemize}
\item $B^t$: original dataset with $b$ points,
\item $C^t$: selected coreset with $c = \sigma b$ points,
\item $\mu=\text{sim}(x,P)$: cosine similarity between the embeddings of point $x$ and fingerprints $P$,
\item $\mu[1] \geq \mu[2] \geq \cdots \geq \mu[b]$: similarity values sorted in descending order,
\item $k = \lfloor \frac{b}{2}\rfloor$: median position,
\item $d(x) = \arccos(\text{sim}(x,P))$: angular distance,
\item $\text{cost}(X) = \frac{1}{|X|}\sum_{x\in X} d(x)$: average angular distance.
\end{itemize}

We first establish four key lemmas:
\begin{lemma}[Selection Interval Bounds] \label{lemma:sib}
For all $x \in C^t$:
\[\mu[k+\lfloor\frac{c}{2}\rfloor] \leq \text{sim}(x,P) \leq \mu[k-\lfloor\frac{c}{2}\rfloor].\]
\end{lemma}

\begin{lemma}[Single Point Change Bound] \label{lemma:spcb}
For any single point change in $C^t$:
\[|\text{cost}_{\text{new}}(C^t) - \text{cost}(C^t)| = \frac{1}{c}|d(x') - d(x)| \leq \frac{\pi}{c},\]
since $d(x) = \arccos(\text{sim}(x,P)) \in [0,\pi].$
\end{lemma}

\begin{lemma}[Similarity Difference Bound] \label{lemma:sdb}
For any $x_1 \in C^t$, $x_2 \in B^t$:
\[|\text{sim}(x_1,P) - \text{sim}(x_2,P)| \leq \max\{|\mu[1] - \mu[k+\lfloor\frac{c}{2}\rfloor]|, |\mu[b] - \mu[k-\lfloor\frac{c}{2}\rfloor]|\}.\]
\end{lemma}

\begin{lemma}[Expected Value Approximation] \label{lemma:eva}
For coreset $C^t$ selected from the middle region of sorted similarity sequence and original batch $B^t$:
\begin{align*}
&\quad |E[\text{cost}(C^t)] - \text{cost}(B^t)| \\
&\leq L\cdot\max\{|\mu[1] - \mu[k+\lfloor\frac{c}{2}\rfloor]|, |\mu[b] - \mu[k-\lfloor\frac{c}{2}\rfloor]|\} = M,  
\end{align*}
where $L$ is the Lipschitz constant of $\arccos$ function and $M$ is constant.
\end{lemma}

\begin{proof}
Let us partition $B^t$ based on similarity values:
\begin{align*}
B_l &= \{x \mid \mu[b] < \text{sim}(x,P) < \mu[k+\lfloor\frac{c}{2}\rfloor]\}, \\
B_m &= \{x \mid \mu[k+\lfloor\frac{c}{2}\rfloor] \leq \text{sim}(x,P) \leq \mu[k-\lfloor\frac{c}{2}\rfloor]\}, \\
B_h &= \{x \mid \mu[k-\lfloor\frac{c}{2}\rfloor] < \text{sim}(x,P) < \mu[1]\}.
\end{align*}
Then, we have:
\[\text{cost}(B^t) = \frac{|B_l|}{b}\cdot\text{cost}(B_l) + \frac{|B_m|}{b}\cdot\text{cost}(B_m) + \frac{|B_h|}{b}\cdot\text{cost}(B_h).\]
Given Lemma~\ref{lemma:sib}, for any point $x \in B_m$:
\[\mu[k+\lfloor\frac{c}{2}\rfloor] \leq \text{sim}(x,P) \leq \mu[k-\lfloor\frac{c}{2}\rfloor],\]
based on the monotonicity of $\arccos$ function, we have:
\[\arccos(\mu[k-\lfloor\frac{c}{2}\rfloor]) \leq d(x) \leq \arccos(\mu[k+\lfloor\frac{c}{2}\rfloor]).\]
Therefore, for any randomly selected point from $B_m$:
\[E[d(x)] \in [\arccos(\mu[k-\lfloor\frac{c}{2}\rfloor]), \arccos(\mu[k+\lfloor\frac{c}{2}\rfloor])].\]
By Lipschitz property of $\arccos$ function: for any $x_1 = \text{sim}(x'_1,P)$ and $x_2 = \text{sim}(x'_2,P)$ where $x_1,x_2 \in [-1,1]$, there exists a Lipschitz constant $L$ such that:
\[ |\arccos(x_1) - \arccos(x_2)| \leq L|x_1 - x_2|.\]
This implies for the expected value:
\begin{align*}
&\quad |E[\text{cost}(B_m)] - \text{cost}(B_m)| \\
&\leq \arccos(\mu[k+\lfloor\frac{c}{2}\rfloor]) - \arccos(\mu[k-\lfloor\frac{c}{2}\rfloor]) \\
&\leq L\cdot|\mu[k+\lfloor\frac{c}{2}\rfloor] - \mu[k-\lfloor\frac{c}{2}\rfloor]|,
\end{align*}
where $L$ is the Lipschitz constant of $\arccos$ function. Similarly, for the differences between expected costs:
\begin{align*}
|E[\text{cost}(B_m)] - \text{cost}(B_l)| \leq L|\mu[b] - \mu[k-\lfloor\frac{c}{2}\rfloor]|,\\
|E[\text{cost}(B_m)] - \text{cost}(B_h)| \leq L|\mu[1] - \mu[k+\lfloor\frac{c}{2}\rfloor]|.   
\end{align*}
Note that:
\begin{align*}
|\mu[k-\lfloor\frac{c}{2}\rfloor] - \mu[k+\lfloor\frac{c}{2}\rfloor]| \leq |\mu[1] - \mu[k+\lfloor\frac{c}{2}\rfloor]|,\\
|\mu[k-\lfloor\frac{c}{2}\rfloor] - \mu[k+\lfloor\frac{c}{2}\rfloor]| \leq |\mu[b] - \mu[k-\lfloor\frac{c}{2}\rfloor]|.  
\end{align*}
By selection strategy:
\[E[\text{cost}(C^t)] = E[\text{cost}(B_m)],\]
we can get:
\begin{align*}
&\quad |E[\text{cost}(C^t)] - \text{cost}(B^t)|  \\
&= |E[\text{cost}(B_m)] - [\frac{|B_l|}{b}\cdot\text{cost}(B_l) + \frac{|B_m|}{b}\cdot\text{cost}(B_m) \\
&\quad + \frac{|B_h|}{b}\cdot\text{cost}(B_h)]| \\
&= |\frac{|B_l|}{b}\cdot E[\text{cost}(B_m)] - \frac{|B_l|}{b}\cdot\text{cost}(B_l)| \\
&\quad + |\frac{|B_m|}{b}\cdot E[\text{cost}(B_m)] - \frac{|B_m|}{b}\cdot\text{cost}(B_m)| \\
&\quad + |\frac{|B_h|}{b}\cdot E[\text{cost}(B_m)] - \frac{|B_h|}{b}\cdot\text{cost}(B_h)| \\
&\leq \frac{|B_l|}{b}\cdot L|\mu[b] - \mu[k-\lfloor\frac{c}{2}\rfloor]| \\
&\quad + \frac{|B_m|}{b}\cdot L|\mu[k-\lfloor\frac{c}{2}\rfloor] - \mu[k+\lfloor\frac{c}{2}\rfloor]| \\
&\quad + \frac{|B_h|}{b}\cdot L|\mu[1] - \mu[k+\lfloor\frac{c}{2}\rfloor]| \\
&\leq \frac{|B_l|}{b}\cdot L\cdot\max\{|\mu[1] - \mu[k+\lfloor\frac{c}{2}\rfloor]|, |\mu[b] - \mu[k-\lfloor\frac{c}{2}\rfloor]|\} \\
&\quad + \frac{|B_m|}{b}\cdot L\cdot\max\{|\mu[1] - \mu[k+\lfloor\frac{c}{2}\rfloor]|, |\mu[b] - \mu[k-\lfloor\frac{c}{2}\rfloor]|\} \\
&\quad + \frac{|B_h|}{b}\cdot L\cdot\max\{|\mu[1] - \mu[k+\lfloor\frac{c}{2}\rfloor]|, |\mu[b] - \mu[k-\lfloor\frac{c}{2}\rfloor]|\} \\
&= L\cdot\max\{|\mu[1] - \mu[k+\lfloor\frac{c}{2}\rfloor]|, |\mu[b] - \mu[k-\lfloor\frac{c}{2}\rfloor]|\} \\
&\quad \cdot(\frac{|B_l|}{b} + \frac{|B_m|}{b} + \frac{|B_h|}{b}) \\
&= L\cdot\max\{|\mu[1] - \mu[k+\lfloor\frac{c}{2}\rfloor]|, |\mu[b] - \mu[k-\lfloor\frac{c}{2}\rfloor]|\} = M \qedhere
\end{align*}   
\end{proof}

Now we proceed with the main proof:

\textbf{Step 1.} We have two bounds: by McDiarmid's inequality from Lemma~\ref{lemma:spcb} where $c=\sigma b$, for any $\xi > 0$: 
\[ \mathbb{P}(|\text{cost}(C^t) - E[\text{cost}(C^t)]| \geq \xi) \leq 2\exp(-2c\xi^2/\pi^2).\]
Then, with probability at least $1-2\exp(-2c\xi^2/\pi^2)$: 
\[|\text{cost}(C^t) - E[\text{cost}(C^t)]| \leq \xi.\]
By Lemma~\ref{lemma:eva}: 
\[|E[\text{cost}(C^t)] - \text{cost}(B^t)| \leq M.\]

\textbf{Step 2.} Let $\xi + M = \varepsilon\cdot\text{cost}(B^t)$, with probability at least $1-2\exp(-2c\xi^2/\pi^2) =1 -2\exp(-2c(\varepsilon\cdot\text{cost}(B^t) - M)^2/\pi^2)$:
\begin{align*}
&\quad |\text{cost}(C^t) - \text{cost}(B^t)| \\
&= |\text{cost}(C^t) - E[\text{cost}(C^t)] + E[\text{cost}(C^t)] - \text{cost}(B^t)| \\
&\leq |\text{cost}(C^t) - E[\text{cost}(C^t)]| + |E[\text{cost}(C^t)] - \text{cost}(B^t)| \\
&\leq \xi + M \\
&= \varepsilon\cdot\text{cost}(B^t).
\end{align*}    

\textbf{Step 3.} Setting this probability to be at least $1-\delta$:
\[1 - 2\exp(-2c(\varepsilon\cdot\text{cost}(B^t) - M)^2/\pi^2) = 1 - \delta\]

\textbf{Step 4.} Solving for $\varepsilon$:
\begin{align*}
\varepsilon &= M/\text{cost}(B^t) + (\pi/\text{cost}(B^t))\sqrt{-\ln(\delta/2)/(2c)}
\end{align*}   
Since:
\begin{itemize}
\item $M = L\cdot\max\{|\mu[1] - \mu[k+\lfloor\frac{c}{2}\rfloor]|, |\mu[b] - \mu[k-\lfloor\frac{c}{2}\rfloor]|\}$ is constant,
\item $0 < \text{cost}(B^t) \leq \pi$,
\item $c = \sigma b$,
\item $-\ln(\delta/2) = \mathcal{O}(\log(1/\delta))$,
\end{itemize}
we have:
\[\varepsilon = \mathcal{O}(\sqrt{\log(1/\delta)/(\sigma b)}).\]

\textbf{Step 5.} Therefore, with probability $\geq 1-\delta$ and $\varepsilon = \mathcal{O}(\sqrt{\log(1/\delta)/(\sigma b)})$:
\begin{align*}
&|\text{cost}(C^t) - \text{cost}(B^t)| \leq \varepsilon\cdot\text{cost}(B^t)\iff  \\
&-\varepsilon\cdot\text{cost}(B^t) \leq \text{cost}(C^t) - \text{cost}(B^t) \leq \varepsilon\cdot\text{cost}(B^t) \iff \\
&(1-\varepsilon)\text{cost}(B^t) \leq \text{cost}(C^t) \leq (1+\varepsilon)\text{cost}(B^t). \qedhere
\end{align*}

\end{proof}

\subsection{Proof of Buffer Update Representativeness Guarantee} \label{appendix:proof_burg}

\begin{theorem}(Buffer Update Representativeness Guarantee)
With probability at least $1-\delta$, the distribution $P_M$ obtained from the buffer update satisfies:
\[
D(P_M, P_t)\leq \varepsilon,
\]
where $D(\cdot,\cdot)$ denotes the Maximum Mean Discrepancy (MMD) between distributions with kernel function $k(x,y) = \text{sim}(x,P)\text{sim}(y,P)$, $P_M$ is the distribution of buffer data, $P_t$ is the distribution of all seen data until time $t$, and $\varepsilon = \mathcal{O}((m\ln(1/\delta))^{1/4})$ with $m$ being the buffer size.
\end{theorem}

\begin{proof}
Let us first define the notations:
\begin{itemize}
\item $M^t$: updated buffer with $m$ points,
\item $P_M$: distribution of buffer data,
\item $P_t$: distribution of all seen data until time $t$,
\item $k(x,y) = \text{sim}(x,P)\text{sim}(y,P)$: kernel function, where $\text{sim}(\cdot)$ is the cosine similarity,
\item $H_m$: $m$-th harmonic number: $\sum_{i=1}^m1/i$,
\item $w_i = 1 - \frac{\frac{1}{i}}{\sum_{j=1}^m1/j}=1-\frac{1}{jH_m}$: based on the rank probability in Eq.\ref{eq:buf_prob} to get the weight for point $i$,
\item $w_{ij} = (1 - \frac{1}{iH_m})(1 - \frac{1}{jH_m})$,
\item $\mathbb{E}_{x,x'\sim P_M}[k(x,x')]=\sum_{i,j=1}w_{ij}k(x,x')$: since $x\sim P_M$ is sampled based on the rank probability,
\item $\mathbb{E}_{y,y'\sim P_t}[k(y,y')]=\sum_{i,j=1}\frac{1}{n^2}k(y,y')$: since $x\sim P_M$ is sampled based on the unity probability,
\item $\mathbb{E}_{x\sim P_M,y\sim P_t}[k(x,y)]=\sum_{i,j=1}w_i\frac{1}{n}k(x,y)$,
\item $\text{MMD}^2(P_M, P_t) = \mathbb{E}_{x,x'\sim P_M}[k(x,x')] + \mathbb{E}_{y,y'\sim P_t}[k(y,y')] - 2\mathbb{E}_{x\sim P_M,y\sim P_t}[k(x,y)]$.
\end{itemize}

\begin{lemma}[$\text{MMD}^2$ Change Bound] \label{lemma:mcb}
When changing the $i$-th sample in the buffer from $x_i$ to $x'_i$, the change in $\text{MMD}^2$ satisfies:
\begin{equation}
|\Delta \text{MMD}^2| \leq 13
\end{equation}
\end{lemma}

\begin{proof}
We analyze the change in $\text{MMD}^2$ in 4 steps:

\textbf{Step 1}: This step is to analyze $\text{MMD}^2$ changes. Given the $\text{MMD}^2$:
\begin{align*}
    \text{MMD}^2(P_M, P_t) &= \mathbb{E}_{x_1,x_2\sim P_M}[k(x_1,x_2)] + \mathbb{E}_{y_1,y_2\sim P_t}[k(y_1,y_2)] \\
    &\quad - 2\mathbb{E}_{x\sim P_M,y\sim P_t}[k(x,y)] \\
    &= \sum_{i,j=1}^m w_{ij}k(x_i,x_j) + \mathbb{E}_{y_1,y_2\sim P_t}[k(y_1,y_2)] \\
    &\quad - 2\sum_{i=1}^m w_i\mathbb{E}_{y\sim P_t}[k(x_i,y)].
\end{align*}
When $x_i$ in $P_M$ changes to $x'_i$, the change in $\text{MMD}^2$:
\begin{align*}
    |\Delta\text{MMD}^2| &\leq |\sum_{i,j=1}^m w'_{ij}k'(x_i,x_j) - \sum_{i,j=1}^m w_{ij}k(x_i,x_j)|\\
    &\quad + 2|\sum_{i=1}^m w'_i\mathbb{E}_{y\sim P_t}[k'(x_i,y)] - \sum_{i'=1}^m w_i\mathbb{E}_{y\sim P_t}[k(x_i,y)]| \\
    &= \Delta(\text{first term}) + (\Delta\text{third term}),
\end{align*}
specifically:
\begin{align*}
(\text{first term})' &= w'_{ii}k(x'_i, x'_i) \tag{self term} \\
&\quad + \sum_{j \neq i}^m w'_i w_j k(x'_i, x_j) \tag{$x'_i$ as the 1-st sample} \\
&\quad + \sum_{j \neq i}^m w_j w'_i k(x_j, x'_i) \tag{$x'_i$ as the 2-nd sample} \\
&\quad + \sum_{p \neq i, q \neq i}^m w_{pq}k(x_p, x_q) \tag{has no $x_i$}.\\
(\text{third term})'& = \sum_{i=1}^m w'_i\mathbb{E}_{y\sim P_t}[k(x'_i,y)].
\end{align*}
Therefore, the change in $\text{MMD}^2$ can be decomposed as:
\begin{align*}
|\Delta \text{MMD}^2| &\leq \Delta(\text{first term}) + \Delta(\text{third term}) \\
&= [w'_{ii}k(x'_i, x'_i) - w_{ii}k(x_i, x_i)] \\
&\quad + 2[w'_i\sum_{j \neq i}^m w_j k(x'_i, x_j) - w_i\sum_{j \neq i}^m w_j k(x_i, x_j)] \\
&\quad + 2[w'_i\mathbb{E}_{y\sim P_t}[k(x'_i, y)] - w_i\mathbb{E}_{y\sim P_t}[k(x_i, y)]]\\
&= \text{Self-term change} + 2(\text{Cross-term change}) \\
&\quad + \text{Expectation-term change}.\\
\end{align*}

\textbf{Step 2}: This step is to analyze weight changes. When a sample changes, its rank may also change. Then:
\begin{align*}
|w'_i - w_i| &= \left|\frac{1}{iH_m} - \frac{1}{(i+\Delta r)H_m}\right| \\
&\leq \frac{\Delta r}{i(i+\Delta r)H_m} \\
&\leq \frac{1}{iH_m}.
\end{align*}

\textbf{Step 3}: Bound the change of each term.\\
\textbf{1) Self-term change}:
\begin{align*}
&\quad |w'_{ii} k(x'_i, x'_i) - w_{ii} k(x_i, x_i)|\\
&= |(1 - \frac{1}{iH_m})^2 \text{sim}^2(x'_i, P) - (1 - \frac{1}{iH_m})^2 \text{sim}^2(x_i, P)|\\
&= |(1 - \frac{1}{iH_m})^2| \cdot |\text{sim}^2(x'_i, P) - \text{sim}^2(x_i, P)| \\
&= |1 - \frac{2}{iH_m} + \frac{1}{i^2H_m^2}| \cdot |\text{sim}^2(x'_i, P) - \text{sim}^2(x_i, P)|.
\end{align*}
Using the inequality $\frac{1}{i^2H_m^2} \leq \frac{1}{iH_m}$ (since $H_m \geq 1$ for $i \geq 1$):
\begin{align*}
|1 - \frac{2}{iH_m} + \frac{1}{i^2H_m^2}| &\leq |1 - \frac{2}{iH_m} + \frac{1}{iH_m}| = |1 - \frac{1}{iH_m}|.
\end{align*}
Therefore:
\begin{align*}
&|w'_{ii} k(x'_i, x'_i) - w_{ii} k(x_i, x_i)| \notag \\
&\leq |1 - \frac{1}{iH_m}| \cdot |\text{sim}^2(x'_i, P) - \text{sim}^2(x_i, P)|.
\end{align*}
Since $\frac{1}{iH_m}$ is positive, $1 - \frac{1}{iH_m} < 1$, and $\text{sim}^2(x'_i, P)$, $\text{sim}^2(x_i, P)$ are in $[0, 1]$, their absolute difference is at most 1:
\begin{align*}
|w'_{ii} k(x'_i, x'_i) - w_{ii} k(x_i, x_i)| \leq 1.
\end{align*}
\textbf{2) Cross-term change:}
\begin{align*}
    &\quad |w'_i \sum_{j\neq i}^m w_j k(x'_i,x_j) - w_i \sum_{j\neq i}^m w_j k(x_i,x_j)| \\
    &= |w'_i \sum_{j\neq i}^m w_j k(x'_i,x_j) - w_i \sum_{j\neq i}^m w_j k(x'_i,x_j) \\
    &\quad + w_i \sum_{j\neq i}^m w_j k(x'_i,x_j) - w_i \sum_{j\neq i}^m w_j k(x_i,x_j)| \\
    &\leq |(w'_i - w_i) \sum_{j\neq i}^m w_j k(x'_i,x_j)| + |w_i \sum_{j\neq i}^m w_j (k(x'_i,x_j) - k(x_i,x_j))| \\
    &\leq |w'_i - w_i| |\sum_{j\neq i}^m w_j k(x'_i,x_j)| + |w_i| |\sum_{j\neq i}^m w_j (k(x'_i,x_j) - k(x_i,x_j))| 
\end{align*}
To solve this, we have:
\begin{itemize}
    \item $|w'_i - w_i| \leq 1/(iH_m)$,
    \item $|k(x'_i, x_j)| = |sim(x'_i,P)sim(x_j,P)| \leq 1$,
    \item $|w_i| = 1/(iH_m)$,
    \item $|k(x'_i, x_j) - k(x_i, x_j)| = |sim(x'_i,P) - sim(x_i,P)||sim(x_j,P)|\leq |sim(x',P) - sim(x,P)|\leq 2$.
\end{itemize}
Therefore:
\begin{align*}
&\quad |w'_i - w_i| |\sum_{j\neq i}^m w_j k(x'_i,x_j)| + |w_i| |\sum_{j\neq i}^m w_j (k(x'_i,x_j) - k(x_i,x_j))| \\
&\leq \frac{1}{iH_m} |\sum_{j\neq i}^m \frac{1}{jH_m}\cdot 1| + \frac{1}{iH_m} |\sum_{j\neq i}^m \frac{1}{jH_m}\cdot 2| \\
&\leq \frac{1}{iH_m} \cdot \frac{H_m}{H_m} + \frac{1}{iH_m} \cdot \frac{2 H_m}{H_m} \\
&= \frac{3}{iH_m} \leq \frac{3}{H_m}.
\end{align*}
\textbf{3) Expectation term change:}
\begin{align*}
&\quad 2|w'_i\mathbb{E}_{y\sim P_t}[k(x'_i, y)] - w_i\mathbb{E}_{y\sim P_t}[k(x_i, y)]| \\
&= 2|w'_i\mathbb{E}_{y\sim P_t}[k(x'_i, y)] - w_i\mathbb{E}_{y\sim P_t}[k(x'_i, y)] \\ 
&\quad + w_i\mathbb{E}_{y\sim P_t}[k(x'_i, y)] - w_i\mathbb{E}_{y\sim P_t}[k(x_i, y)]| \\
&\leq 2|w'_i - w_i||\mathbb{E}_{y\sim P_t}[k(x'_i, y)]|  + 2|w_i||\mathbb{E}_{y\sim P_t}[k(x'_i, y) - k(x_i, y)]| 
\end{align*}
To solve this, we have:
\begin{itemize}
    \item $k(x'_i, y)=sim(x'_i,P)sim(y,P) \leq 1$,
    \item $\mathbb{E}_{y\sim P_t}[k(x'_i, y)] \leq \max(k(x'_i, y))\leq 1$,
    \item \parbox{\linewidth}{
    \begin{align*}
        &\quad |E_{y\sim P_t}[k(x'_i,y) - k(x_i,y)]| \\
        &= |E_{y\sim P_t}[sim(x'_i,P)sim(y,P) - sim(x_i,P)sim(y,P)]| \\
        &= |E_{y\sim P_t}[(sim(x'_i,P) - sim(x_i,P))sim(y,P)]| \\
        &= |sim(x'_i,P) - sim(x_i,P)| |E_{y\sim P_t}[sim(y,P)]| \leq 2.
    \end{align*}
    }
\end{itemize}
Therefore:
\begin{align*}
    &\quad 2|w'_i - w_i||\mathbb{E}_{y\sim P_t}[k(x'_i, y)]| + 2|w_i||\mathbb{E}_{y\sim P_t}[k(x'_i, y) - k(x_i, y)]| \\
    &\leq \frac{2}{iH_m}\cdot 1 + \frac{2}{iH_m}\cdot 2 
    =\frac{6}{H_m}.
\end{align*}

\textbf{Step 4}: Combine all the above bounds:
\begin{align*}
|\Delta \text{MMD}^2| &= \text{Self-term change} + 2(\text{Cross-term change}) \\
                      &\quad + \text{Expectation-term change} \\
                      &\leq 1 + 2\cdot\frac{3}{H_m} + \frac{6}{H_m} \\
                      &= 1+\frac{12}{H_m} \leq 13. \codecomment{\quad// $H_M\geq1$} \qedhere
\end{align*}
\end{proof}
\begin{lemma}[$\text{MMD}^2$ Expectation Bound] \label{lemma:meb}
    For $\mathbb{E}[\text{MMD}^2]$, it satisfies: $\mathbb{E}[\text{MMD}^2] \leq A$, where $A=\frac{\pi^2}{6H_m^2} + 1 - \frac{2}{H_m^2} \sum_{i=1}^m \frac{H_i}{i}$.
\end{lemma}
\begin{proof}
We analyze the expectation bound of $\text{MMD}^2$ in 2 steps:  

\textbf{Step 1}: Analyze the $\text{MMD}^2$ expectation:
\begin{align*}
\mathbb{E}[\text{MMD}^2] &= \mathbb{E}[\sum_{i,j=1}^m w_{ij}k(x_i,x_j)] + \mathbb{E}[\mathbb{E}_{y,y'\sim P_t}[k(y,y')]]  \\
&\quad - 2\mathbb{E}[\sum_{i=1}^m w_i\mathbb{E}_{y\sim P_t}[k(x_i,y)]].
\end{align*}
\textbf{1) First term expectation}:
\begin{align*}
    &\quad \mathbb{E}[\sum_{i=1}^m \sum_{j=1}^m w_i w_j k(x_i, x_j)] \\
    &= \mathbb{E}[\sum_{i=1}^m w_i^2 k(x_i, x_i)] 
      + \mathbb{E}[\sum_{i=1}^m \sum_{j=i+1}^m w_i w_j k(x_i, x_j)]\\ 
   &\quad + \mathbb{E}[\sum_{j=1}^m \sum_{i=j+1}^m w_i w_j k(x_i, x_j)] \\
   &= \mathbb{E}[\sum_{i=1}^m w_i^2 k(x_i, x_i)]  + 2\mathbb{E}[\sum_{i=1}^m \sum_{j=i+1}^m w_i w_j k(x_i, x_j)].
\end{align*}
We can get that:
\begin{align*}
    \mathbb{E}[\sum_{i=1}^m w_i^2 k(x_i, x_i)] 
    &\leq \mathbb{E}[\sum_{i=1}^m (\frac{1}{i H_m})^2] \codecomment{\quad// $k(x_i, x_i) \leq 1$} \\
    &= \mathbb{E}[\frac{1}{H_m^2}\sum_{i=1}^m\frac{1}{i^2}] \\
    &\leq \mathbb{E}[\frac{1}{H_m^2}\frac{\pi^2}{6}] \codecomment{\quad// $\sum_{i=1}^m\frac{1}{i^2} \leq \sum_{i=1}^\infty\frac{1}{i^2} =\frac{\pi^2}{6}$} \\
    &= \frac{\pi^2}{6H_m^2}, \codecomment{\quad// $H_m^2$ is deterministic}
\end{align*}
\begin{align*}
&\quad\mathbb{E}[\sum_{i=1}^m \sum_{j=i+1}^m w_i w_j k(x_i, x_j)] \\
&= \sum_{i=1}^m \sum_{j=i+1}^m w_i w_j \mathbb{E}[k(x_i, x_j)] \codecomment{\quad//$w$ are deterministic}\\
&\leq \sum_{i=1}^m \sum_{j=i+1}^m \frac{1}{iH_m} \cdot \frac{1}{jH_m} \cdot 1 \\
&= \frac{1}{H_m^2} \sum_{i=1}^m \frac{1}{i} \sum_{j=i+1}^m \frac{1}{j} \\
&= \frac{1}{H_m^2} \sum_{i=1}^m \frac{1}{i} (H_m - H_i) \\
&= \frac{1}{H_m^2} \left(\sum_{i=1}^m \frac{H_m}{i} - \sum_{i=1}^m \frac{H_i}{i}\right) \\
&= \frac{1}{H_m^2} (H_m \cdot H_m - \sum_{i=1}^m \frac{H_i}{i}) \\
&= 1 - \frac{1}{H_m^2} \sum_{i=1}^m \frac{H_i}{i}.
\end{align*}
Therefore:
\begin{align*}
    \mathbb{E}[\sum_{i=1}^m \sum_{j=1}^m w_i w_j k(x_i, x_j)]
    &\leq \frac{\pi^2}{6H_m^2} + 2(1 - \frac{1}{H_m^2} \sum_{i=1}^m \frac{H_i}{i}) \\
    &= \frac{\pi^2}{6H_m^2} + 2 - \frac{2}{H_m^2} \sum_{i=1}^m \frac{H_i}{i}.
\end{align*}
\textbf{2) Second term expectation}:
\begin{align*}
\mathbb{E}[\mathbb{E}_{y,y'\sim P_t}[k(y,y')]] 
= \mathbb{E}_{y,y'\sim P_t}[k(y,y')] \leq 1.
\end{align*}
\textbf{3) Third term expectation}:
\begin{align*}
\mathbb{E}[\sum_{i=1}^m w_i\mathbb{E}_{y\sim P_t}[k(x_i,y)]] 
&\leq \mathbb{E}[\sum_{i=1}^m w_i] \\
&= \sum_{i=1}^m\frac{1}{iH_m} \\
&= \frac{1}{H_m} \cdot H_m = 1.
\end{align*}

\textbf{Step 2}: combine these expectation bounds:
\begin{align*}
\mathbb{E}[\text{MMD}^2] &\leq (\frac{\pi^2}{6H_m^2} + 2 - \frac{2}{H_m^2} \sum_{i=1}^m \frac{H_i}{i}) + 1 - (2 \times 1) \\
&= \frac{\pi^2}{6H_m^2} + 1 - \frac{2}{H_m^2} \sum_{i=1}^m \frac{H_i}{i}.
\end{align*}
Let $A=\frac{\pi^2}{6H_m^2} + 1 - \frac{2}{H_m^2} \sum_{i=1}^m \frac{H_i}{i}$ and $\mathbb{E}[\text{MMD}^2] \leq A$.
\end{proof}

Now we proceed with the main proof:

\textbf{Step 1}: From Lemma~\ref{lemma:mcb} that $|\Delta\text{MMD}^2|\leq 13$, we can apply McDiarmid's inequality:
\[
\mathbb{P}(|\text{MMD}^2 - \mathbb{E}[\text{MMD}^2]| \geq \xi) \leq 2exp(-2\xi^2/169m),
\]
where $m$ is the buffer size.
Then, with probability at least $1-2exp(-2t^2/169m)$:
\begin{align*}
    &|\text{MMD}^2 - \mathbb{E}[\text{MMD}^2]| \leq \xi \iff\\
    &\mathbb{E}[\text{MMD}^2] - \xi \leq \text{MMD}^2 \leq \mathbb{E}[\text{MMD}^2] + \xi
\end{align*}
Focus on upper bound, since MMD is non-negative:
\begin{align*}
&\text{MMD}^2 \leq \mathbb{E}[\text{MMD}^2] + \xi \implies\\
&\text{MMD} \leq \sqrt{\mathbb{E}[\text{MMD}^2] + \xi} \leq \sqrt{\mathbb{E}[\text{MMD}^2]} + \sqrt{\xi} \leq \sqrt{A}+\sqrt{\xi},
\end{align*}
where the last inequality follows from $\sqrt{a+b} \leq \sqrt{a} + \sqrt{b}$ for $a,b \geq 0$. \\
Combining this with McDiarmid's inequality above, we have:
\begin{align*}
    &\mathbb{P}(|MMD^2 - E[MMD^2]| \leq \xi) \geq 1-2exp(-2\xi^2/169m) \iff\\
    &\mathbb{P}(\text{MMD} \leq \sqrt{A}+\sqrt{\xi}) \geq 1 - 2exp(-2\xi^2/169m).
\end{align*}

\textbf{Step 2}: Let $\xi' = \sqrt{A} + \sqrt{\xi}$:
\begin{align*}
\sqrt{\xi} = \xi' - \sqrt{A},\\
\xi = (\xi' - \sqrt{A})^2.
\end{align*}
Hence, we have:
\begin{align*}
\mathbb{P}(\text{MMD} \leq \xi') &\geq 1 - 2exp(-2\xi^2/169m)\\
&= 1 - 2exp(-2(\xi' - \sqrt{A})^4/169m)  
\end{align*}

\textbf{Step 3}: Let $\delta = 2\exp(-2(\xi'-\sqrt{A})^4/169)$, then:
\[\mathbb{P}(MMD \leq \varepsilon) \geq 1 - \delta\]

\textbf{Step 4}: Solve for $\xi'$:
\begin{align*}
\delta &= 2\exp(-2(\xi'-\sqrt{A})^4/169m) \\
\ln(\delta/2) &= -2(\varepsilon-\sqrt{A})^4/169m \\
\varepsilon &= \sqrt{A} + (-84.5m\ln(\delta/2))^{1/4}
\end{align*}

\textbf{Step 5}: Note that, since $H_m$ is the harmonic series:
\[
A = \frac{\pi^2}{6H_m^2} + 1 - 2\sum_{i=1}^m\frac{H_i}{iH_m^2} = \mathcal{O}(1).
\]
Therefore:
\[\sqrt{A} = \mathcal{O}(1).\]
For the second term:
\begin{align*}
(-84.5m\ln(\delta/2))^{1/4} &= (84.5m\ln(2) - 84.5m\ln(\delta))^{1/4} \\
&= \mathcal{O}((m\ln(1/\delta))^{1/4}).
\end{align*}

\textbf{Step 6}: Finally, renaming $\xi'$ to $\varepsilon$, we obtain:
\[\mathbb{P}(MMD(P_M, P_t) \leq \varepsilon) \geq 1-\delta,\]
where:
\[\varepsilon = \mathcal{O}(1) + \mathcal{O}((m\ln(1/\delta))^{1/4}) = \mathcal{O}((m\ln(1/\delta))^{1/4}). \qedhere\]
\end{proof}